\documentclass[pdflatex,sn-mathphys-num]{sn-jnl}
\usepackage{graphicx}
\usepackage{amsmath,amssymb,amsfonts}
\usepackage{amsthm}
\usepackage{mathrsfs}
\usepackage[title]{appendix}
\usepackage[dvipsnames]{xcolor}
\usepackage{textcomp}
\usepackage{booktabs}
\usepackage{algorithm}
\usepackage{algorithmicx}
\usepackage{algpseudocode}
\usepackage{bm}
\usepackage{subcaption}

\theoremstyle{thmstyleone}

\theoremstyle{thmstyletwo}

\theoremstyle{thmstylethree}

\usepackage{tikz} 
\usetikzlibrary{shapes.geometric, arrows.meta, positioning, fit, backgrounds}

\usepackage{multirow}
\usepackage{siunitx}
\usepackage{mathabx}

\begin{document}

\title[Dimensionally consistent surrogate modelling through dimensional analysis and harmonic expansions]{{Dimensionally consistent surrogate modelling through dimensional analysis and harmonic expansions}}

\author[1]{\fnm{Ernest Tarrus}}\email{ernesttarrus@gmail.com}

\author[1]{\fnm{Hector Gisbert}}\email{hector.gisbert@universidadeuropea.es}

\affil*[1]{
\orgname{Universidad Europea de Valencia},
\orgdiv{Escuela de Ciencias, Ingeniería y Diseño},
\orgaddress{\street{C/ de Guillem de Castro, 175, Extramurs}, \city{Valencia}, \postcode{46008}, \state{Valencia}, \country{Spain}}\\[0.15cm]
\textbf{Corresponding Author:} Hector Gisbert}

\abstract{
{Dimensional homogeneity is a fundamental constraint on physically meaningful
models, requiring invariance under changes of units. We present a data-driven
method for constructing surrogate models that satisfy this constraint at the
level of the hypothesis class. Starting from a dimension matrix of measured
variables, the method derives Buckingham $\Pi$-groups, constructs admissible
dimensional prefactors, and approximates the remaining dimensionless dependence
using truncated harmonic expansions on normalized invariant domains. Once the
prefactor and dictionary are fixed, the coefficients are obtained from a
regularized linear regression problem. We test the approach on the simple
pendulum, Planck's black-body law, the double-pendulum Lyapunov field, and an
experimental COBE/FIRAS black-body spectrum dataset.
The results show that dimensional constraints improve conditioning, robustness
to noise, and sample efficiency relative to unconstrained baselines, while the
choice of dictionary becomes important in non-periodic or multi-invariant
settings. The learned expressions are explicit and inexpensive to evaluate,
which makes them useful as surrogate models for structured physical problems.
}
}


\maketitle

\section{Introduction}\label{sec:introduction}

{Dimensional analysis plays a central role in the physical sciences. At a
minimum, it provides a consistency criterion: physically meaningful equations
must be dimensionally homogeneous, so that their form is preserved under
changes of units \cite{bridgman1922,barenblatt2003}. More importantly, it
reveals the internal structure of physical models. By exploiting scaling
relations among variables, dimensional analysis reduces the effective number of
degrees of freedom and identifies the dimensionless combinations that govern
the remaining behavior, as formalized by the Buckingham $\Pi$ theorem
\cite{buckingham1914,barenblatt1996,sedov1993,barenblatt2003}. This result is
also known as the Vaschy--Buckingham theorem, since Vaschy stated an equivalent
formulation in 1892 \cite{vaschy1892}; we follow the standard convention and
refer to it as the Buckingham $\Pi$ theorem throughout. The resulting
invariants provide natural coordinates for the problem and clarify which
combinations of variables may influence the observable of interest.
}

The growing interest in data-driven modelling of physical systems has renewed
the relevance of such structural constraints. Symbolic regression seeks
explicit analytic expressions directly from data \cite{schmidt2009,udrescu2020},
while sparse-library approaches identify compact models from prescribed
candidate terms \cite{brunton2016,rudy2017}. More broadly, physics-informed
learning introduces prior physical knowledge into the learning process in order
to improve generalization, robustness, and interpretability \cite{raissi2019}.
In all these settings, one basic requirement is often overlooked or only weakly
enforced: the learned model should remain meaningful under arbitrary changes of
units. A predictor that depends on whether length is expressed in meters or
centimeters is not physically admissible. This suggests that dimensional
consistency should not be treated merely as an a posteriori check, but rather
imposed directly at the level of the hypothesis class.

Recent work has begun to incorporate dimensional structure more explicitly into
machine-learning pipelines. Villar et al.\ formulate learning with physical
quantities as a problem of exact units equivariance, showing how dimensional
analysis can be used to construct invariant representations before inference
\cite{villar2023units}. Bakarji et al.\ develop dimensionally consistent
learning with Buckingham $\Pi$, including constrained fitting and data-driven
identification of dimensionless groups and scaling parameters
\cite{bakarji2022buckingham}. Xie et al.\ propose a framework for discovering
dominant dimensionless numbers and governing relations from limited
measurements \cite{xie2022dimensionless}. Related ideas have also been explored
in data-driven dimensional analysis for complex flow systems, where the goal is
to identify informative nondimensional groups directly from simulation or
experimental data \cite{jofre2020multiphase}. Taken together, these works show
that dimensional structure is not only a consistency requirement, but also a
useful inductive bias for modelling physical relations from data.

In this work, we construct surrogate models from data under explicit
dimensional constraints. Starting from the dimensional structure of the
measured variables, we form independent Buckingham invariants and identify the
admissible dimensional scalings compatible with the target quantity. The
predictor is written as a dimensional prefactor multiplied by a dimensionless
modulation on invariant space, and the modulation is approximated with a
truncated harmonic dictionary. With this choice, fitting the coefficients
reduces to regularized linear regression, for example Ridge, Lasso, or Elastic
Net \cite{tikhonov1977,hoerl1970,tibshirani1996,zou2005}.

{Our approach is related to this previous literature, and several of its
individual ingredients are classical. Buckingham dimensional analysis,
monomial dimensional prefactors, harmonic/Fourier approximations, and Ridge
regularization are all well-established tools. The contribution of the present
work is to use these components as a fixed hypothesis class for surrogate
modelling: an admissible dimensional prefactor is selected from the affine space
$A\bm{\beta}=\mathbf{b}$, while the remaining dimensionless response is
approximated by a finite dictionary on Buckingham invariant space.

This differs from previous dimensionally aware learning methods in scope and
emphasis. Existing approaches often focus on discovering dimensionless groups
from data, embedding dimensional constraints into flexible black-box
regressors, or enforcing units equivariance in broad model classes
\cite{villar2023units,bakarji2022buckingham,xie2022dimensionless}. Here, the
admissible dimensional structure and the approximation space are specified
before fitting. The model is therefore less general than a symbolic discovery
engine, but its assumptions and fitted coefficients are directly inspectable.
In this sense, our framework is also distinct from symbolic-regression systems
such as AI Feynman, which search over broader symbolic expression spaces rather
than working within a structured approximation class determined a priori by
dimensional admissibility \cite{udrescu2020}.}

{The main contributions of this work are therefore the following. First, we
formulate dimensionally constrained surrogate modelling as a unit-equivariant
regression problem determined by the dimension matrix of the variables. Second,
we construct an explicit prefactor-times-invariants hypothesis class in which
the admissible dimensional scaling is separated from the residual
dimensionless response. Third, we study finite dictionaries on invariant space,
including one-dimensional, phase-combination, and separable tensor-product
forms, and discuss when the associated periodic continuation is appropriate.
Fourth, we evaluate the resulting models against unconstrained and
dimensionally aware baselines, including alternative non-periodic basis
choices, on synthetic and experimental benchmark data.}

The benchmarks are chosen to separate four effects: recovery of a dimensional
skeleton, approximation of a nontrivial dimensionless modulation, validation on
measured data, and dictionary design in a multi-invariant setting. Across these cases, the experiments track how dimensional constraints affect
conditioning, robustness, and sample efficiency, and how dictionary geometry
matters once the invariant space is multidimensional.

The remainder of the paper is organized as follows.
Section~\ref{sec:mathematical_framework} introduces the dimensional
formulation of the learning problem. Section~\ref{sec:learning_physical_laws}
develops the unit-consistent hypothesis class, including admissible
prefactors, invariant coordinates, and harmonic dictionaries.
Section~\ref{sec:algorithm} summarizes the full learning pipeline.
Sections~\ref{sec:experiments}, \ref{sec:results}, and \ref{sec:discussion}
present the benchmarks and discuss the numerical results. Finally,
Section~\ref{sec:conclusion} concludes with the main implications and possible
extensions of the framework.

\section{Dimensional structure of the learning problem}
\label{sec:mathematical_framework}

We begin by formalizing the dimensional structure that constrains the admissible form of the target response. The aim of this section is to introduce the dimension matrix, the notion of unit equivariance, and the reduction to dimensionless invariant coordinates provided by the Buckingham $\Pi$ theorem.

\subsection{Variables, dimensions, and the dimension matrix}

Consider a physical system described by measured variables $\mathbf{x}=(x_1,\dots,x_N)$ and a target quantity $y$, related through an unknown response function
\begin{equation}
y=f(x_1,\dots,x_N)=f(\mathbf{x}).
\label{eq:unknown_law}
\end{equation}
In practice, this relation may also depend on additional quantities that are unobserved, externally controlled, or effectively fixed over the dataset under consideration. We do not model such variables explicitly and assume that, on the domain of interest, their effect is either negligible or approximately constant.

Let $\mathcal{B}=\{D_1,\dots,D_R\}$ be a basis of fundamental dimensions, such as mass, length, and time. Each input variable $x_j$ is assigned a dimensional signature
\begin{align}
[x_j]=\prod_{r=1}^{R} (D_r)^{a_{rj}},
\qquad a_{rj}\in\mathbb{Q},
\qquad j=1,\dots,N,
\end{align}
and the target variable satisfies
\begin{align}
[y]=\prod_{r=1}^{R} (D_r)^{b_r},
\qquad b_r\in\mathbb{Q}.
\end{align}
The exponents of the input variables are collected in the \emph{dimension matrix}
\begin{equation}
A=(a_{rj})\in\mathbb{Q}^{R\times N},
\label{eq:dimension_matrix}
\end{equation}
whose $j$-th column encodes the dimensional signature of $x_j$. This representation allows the dimensional structure of the problem to be handled algebraically. In particular, linear relations among the columns of $A$ correspond to dimensionless combinations of the inputs, while relations involving the target signature $\mathbf{b}=(b_1,\dots,b_R)^\top$ determine the admissible dimensional scalings of the response.

\subsection{Unit equivariance}

A physically meaningful model must preserve its form under changes of units. This requirement is naturally expressed as an equivariance condition under the scaling group $G=(\mathbb{R}_{>0})^R$. For $\lambda=(\lambda_1,\dots,\lambda_R)\in G$, the induced action on each input variable $x_j$ is
\begin{equation}
x_j \mapsto \left(\prod_{r=1}^R (\lambda_r)^{a_{rj}}\right)x_j,
\qquad j=1,\dots,N,
\label{eq:input_scaling}
\end{equation}
while the target transforms as
\begin{equation}
y \mapsto \left(\prod_{r=1}^R (\lambda_r)^{b_r}\right)y.
\label{eq:target_scaling}
\end{equation}
Accordingly, the response function~\eqref{eq:unknown_law} is physically admissible only if it satisfies
\begin{equation}
f\left(\left(\prod_{r=1}^R (\lambda_r)^{a_{r1}}\right)x_1,\,\cdots\,,\left(\prod_{r=1}^R (\lambda_r)^{a_{rN}}\right)x_N\right)
=
\left(\prod_{r=1}^R (\lambda_r)^{b_r}\right)f(\mathbf{x})~.
\label{eq:unit-equivariance}
\end{equation}

Equation~\eqref{eq:unit-equivariance} is the mathematical expression of dimensional homogeneity. It states that the learned response must transform exactly as the target quantity under any admissible rescaling of the fundamental units. A formal equivariance theorem
for this class of monomial-times-invariant predictors is established in the
broader machine-learning setting by~\cite{villar2023units}; the differential
form of the constraint specific to our setup is derived in
Appendix~\ref{app:pde_proof}.

\subsection{Dimensionless invariants and admissible prefactors}

Dimensional analysis implies that the unit-equivariance condition strongly restricts the admissible form of the response. A monomial
\begin{equation}
\pi(\mathbf{x})=\prod_{j=1}^{N} (x_j)^{\gamma_j}
\label{eq:pi_monomial_general}
\end{equation}
is dimensionless if and only if its exponent vector $\bm{\gamma}\in\mathbb{Q}^N$ satisfies
\begin{equation}
A\bm{\gamma}=\bm{0}~.
\label{eq:kernel_condition}
\end{equation}
Thus, dimensionless combinations of the inputs are in one-to-one correspondence with the kernel of the dimension matrix. If $\{\bm{\gamma}^{(m)}\}_{m=1}^{N_\Pi}$ is a basis of $\ker(A)$, then the associated Buckingham invariants are
\begin{equation}
\pi_m(\mathbf{x})=\prod_{j=1}^{N} (x_j)^{\gamma_j^{(m)}},
\qquad m=1,\dots,N_\Pi,
\label{eq:pi_basis}
\end{equation}
with $N_\Pi=\dim\ker(A)=N-\operatorname{rank}(A)$. In parallel, admissible dimensional prefactors are obtained by solving
\begin{equation}
A\bm{\beta}=\mathbf{b}~,
\label{eq:prefactor_affine_constraint}
\end{equation}
where $\mathbf{b}=(b_1,\dots,b_R)^\top$ is the dimensional signature of the target.
We denote the solution set by $\mathcal{S}_b = \{\bm{\beta}\in\mathbb{Q}^N \mid A\bm{\beta}=\mathbf{b}\}$. Any solution $\bm{\beta}\in\mathcal{S}_b$ defines a monomial
\begin{equation}
P_{\bm{\beta}}(\mathbf{x})=\prod_{j=1}^{N} x_j^{\beta_j}
\label{eq:prefactor_monomial}
\end{equation}
with the same physical dimensions as $y$. If $\bm{\beta}_0$ is one particular solution, then every other solution has the form
\begin{equation}
\bm{\beta}=\bm{\beta}_0+\mathbf{v},
\qquad \mathbf{v}\in\ker(A)~,
\end{equation}
{where $\mathcal{S}_b$ is an affine subspace parallel to $\ker(A)$, with dimension $N_\Pi=\dim\ker(A)$. The vector $\mathbf{v}$ should therefore be understood as an arbitrary element of this kernel, not as a single fixed displacement,
so different admissible prefactors differ only by multiplication of a dimensionless monomial.}\footnote{In the benchmark sections, when the free parameters of $\mathcal{S}_b$ reduce to a single scalar or a pair, we write $P_\alpha$ or $P_{\eta,\xi}$ as convenient specializations of the general monomial $P_{\bm{\beta}}$.}

Whenever there exists $\bm{\beta}\in\mathbb{Q}^N$ such that $A\bm{\beta}=\mathbf{b}$, the Buckingham $\Pi$ theorem implies that any dimensionally homogeneous response can be written as a dimensional prefactor times a dimensionless function of the invariants. In the finite model class used here, this motivates predictors of the form
\begin{equation}
{
y=\sum_{\bm{\beta}\in\mathcal{T}} 
P_{\bm{\beta}}(\mathbf{x})\,
\Phi_{\bm{\beta}}\bigl(\pi_1(\mathbf{x}),\dots,\pi_{N_\Pi}(\mathbf{x})\bigr),}
\label{eq:prefactor_times_modulation}
\end{equation}
{where $\mathcal{T}\subset\mathcal{S}_b$ denotes a finite set of admissible
candidate exponent vectors selected from the affine solution space of
$A\bm{\beta}=\mathbf{b}$. Thus, the sum is not taken over the full continuous
solution space, but only over the finite prefactor dictionary used in the
regression model. Here, $P_{\bm{\beta}}$ carries the physical dimensions of the
target $y$ and $\Phi_{\bm{\beta}}$ is dimensionless. Different admissible choices
of $\bm{\beta}$ correspond to equivalent factorizations of the same response, since
any change in the prefactor can be absorbed into the dimensionless modulation
$\Phi_{\bm{\beta}}$.
}

In the experiments reported here, each benchmark employs a single monomial
prefactor, so the sum contains one term:
$$
y=P_{\bm{\beta}_0}(\mathbf{x})\,\Phi_{\bm{\beta}_0}\bigl(\pi_1(\mathbf{x}),\dots,\pi_{N_\Pi}(\mathbf{x})\bigr).
$$

In the next section, we turn this structural representation into a concrete hypothesis class by specifying how the dimensionless modulation is approximated and how the resulting model is fitted from observations.

\section{Dimensionally consistent hypothesis class}
\label{sec:learning_physical_laws}

The dimensional analysis developed in Section~\ref{sec:mathematical_framework} fixes the admissible dimensional structure of the predictor, but not the form of the remaining dimensionless dependence. We now specify this dependence by writing the target response as a dimensional prefactor $P_{\bm{\beta}}$ multiplied by a modulation $\Phi_{\bm{\beta}}$ on invariant space, and then approximating that modulation with harmonic dictionaries.

\subsection{Prefactor-times-invariants representation}

Let $\{\pi_m\}_{m=1}^{N_\Pi}$ be a basis of independent Buckingham invariants and let $P_{\bm{\beta}}$ be any admissible prefactor satisfying $[P_{\bm{\beta}}]=[y]$. We seek predictors of the form
\begin{equation}
g(\mathbf{x})=\sum_{\bm{\beta}\in\mathcal{T}}  P_{\bm{\beta}}(\mathbf{x})\, \Phi_{\bm{\beta}}\bigl(\pi_1(\mathbf{x}),\dots,\pi_{N_\Pi}(\mathbf{x})\bigr),
\label{eq:model_prefactor_phi}
\end{equation}
where $\Phi_{\bm{\beta}}$ are dimensionless modulation functions to be learned from data. This representation directly enforces unit equivariance. Indeed, the prefactor carries the full dimensional signature of the target, whereas the invariants remain unchanged under unit transformations. As a result, any predictor of the form \eqref{eq:model_prefactor_phi} satisfies the scaling law in Eq.~\eqref{eq:unit-equivariance} by construction.

\subsection{Harmonic approximation on invariant space}

The remaining task is to approximate the dimensionless modulation functions $\Phi_{\bm{\beta}}$ on the invariant coordinates. Since the Buckingham invariants need not be periodic or bounded on their raw scale, we first map them to a normalized compact domain. For each invariant $\pi_m$, we define
\begin{equation}
u_m(\pi_m)
=
2\pi\,
\left(\frac{\pi_m-\pi_m^{\min}}{\pi_m^{\max}-\pi_m^{\min}}\right),
\qquad m=1,\dots,N_\Pi,
\label{eq:invariant_rescaling}
\end{equation}
{If $\pi_m^{\max}=\pi_m^{\min}$ for a given invariant, the corresponding coordinate has zero empirical variation and the affine normalization is not defined. In that degenerate case, we set $u_m=0$ and remove all non-constant harmonic features associated with that invariant. This reflects the fact that no dependence on that coordinate can be inferred from the available data.}
For notational convenience, the dependence of $\pi_m$ on $\mathbf{x}$ is left implicit, so we write $u_m(\pi_m)$ in place of $u_m(\pi_m(\mathbf{x}))$. Furthermore, we write $\Phi(u_1,\dots,u_{N_\Pi})\equiv\Phi_{\bm{\beta}}(u_1(\pi_1),\dots,u_{N_\Pi}(\pi_{N_\Pi}))$ for the modulation expressed in normalized coordinates; this is the same dimensionless factor as in \eqref{eq:prefactor_times_modulation}, with the invariants $\pi_m$ replaced by their affinely rescaled counterparts $u_m$.

Here, $\pi_m^{\min}$ and $\pi_m^{\max}$ are estimated from the dataset or prescribed from prior physical constraints. This maps the observed invariant domain into $[0,2\pi]^{N_\Pi}$. We then approximate $\Phi_{\bm{\beta}}$ by a truncated harmonic expansion in the rescaled variables $u=(u_1,\dots,u_{N_\Pi})$.

{The harmonic dictionary is a modelling choice, not a consequence of
dimensional consistency. It is natural when the invariants have angular or
phase-like character, and the truncation order has a simple spectral
interpretation. It also leaves a standard regularized linear regression problem
for the coefficients.

Other basis choices are possible within the same dimensional formulation. For
bounded non-periodic invariant domains, polynomial bases such as Chebyshev or
Legendre features may be more appropriate. For localized or non-smooth
modulations, splines or radial basis functions could also be used. These
alternatives would keep the same prefactor and Buckingham invariants, changing
only the dictionary used for the residual dimensionless modulation.}

{ The affine embedding in Eq.~\eqref{eq:invariant_rescaling} introduces a periodic continuation of the learned dimensionless modulation. For example, in the double-pendulum benchmark the relevant invariant coordinates are the angles $\theta_1,\theta_2\in[-\pi,\pi]$, so identifying the endpoints is consistent with the geometry of the problem.

By contrast, in the black-body benchmark the invariant $\pi_1=k_B T/(h\nu)$ is not periodic. In that case, the harmonic dictionary should be understood as a finite spectral surrogate on the sampled interval, not as a physically periodic representation of the Planck modulation. If the endpoint values of the modulation differ substantially, the periodic extension can introduce boundary artefacts and slower convergence, analogous to Gibbs-type effects. For this reason, we explicitly compare the harmonic dictionary with a Chebyshev polynomial dictionary in the black-body benchmark. Chebyshev features provide a natural non-periodic basis on a bounded interval and therefore serve as a diagnostic of whether the harmonic continuation is limiting the approximation.
}

\subsubsection{One-dimensional case}
If $N_\Pi=1$, we use
\begin{equation}
\Phi(u_1) \approx w_0 + \sum_{n=1}^{N_f} \Bigl[ w_n^{c}\cos(n u_1) + w_n^{s}\sin(n u_1) \Bigr],
\label{eq:fourier_1d_model}
\end{equation}
where $N_f$ is the maximum retained frequency. In the multi-invariant case, the truncation orders per coordinate are written $\{N_f^{(m)}\}_{m=1}^{N_\Pi}$; when a uniform order is used across all coordinates it is abbreviated to the scalar $N_f$ (i.e.\ $N_f^{(m)}=N_f$ for all $m$).

\subsubsection{Multi-invariant phase-combination dictionary}

For $N_\Pi>1$, a direct multidimensional extension of the one-dimensional harmonic expansion is obtained by introducing a non-negative integer multi-index
\begin{equation}
\mathbf{n}=(n_1,\dots,n_{N_\Pi})\in(\mathbb{Z}_{\geq0})^{N_\Pi},
\end{equation}
and defining the corresponding phase through the dot product between the multi-index vector and the invariant-coordinate vector $\mathbf{n}\cdot u = n_1u_1+\cdots+n_{N_\Pi}u_{N_\Pi}$. The dimensionless modulation is then approximated as
\begin{equation}
\Phi(u) \approx w_0 + \sum_{\mathbf{n}\in\mathcal{I}_{\mathbf{n}}} \Bigl[ w_{\mathbf{n}}^{c}\cos(\mathbf{n}\cdot u) + w_{\mathbf{n}}^{s}\sin(\mathbf{n}\cdot u) \Bigr],
\label{eq:fourier_phase_dictionary}
\end{equation}
where $\mathcal{I}_{\mathbf{n}}\subset(\mathbb{Z}_{\geq0})^{N_\Pi}\setminus\{\mathbf{0}\}$ is a finite truncation set. The main feature of this basis is that all invariant coordinates are coupled through a single phase. This makes the representation compact and natural, but it may also limit flexibility when the target function exhibits strongly directional or anisotropic structure. In such settings, the approximation may benefit from a separable tensor-product basis, which resolves the behavior along different invariant directions more explicitly.

\subsubsection{Separable tensor-product dictionary}
In multi-invariant problems, especially when the response surface is anisotropic, it is often preferable to use a separable tensor-product basis. For each coordinate $u_m$, define the one-dimensional basis set $\mathcal{F}_m = \{\psi^{(m)}_k\}_{k=0}^{2N_f^{(m)}}$ ordered as:
\begin{equation}
\psi^{(m)}_{0}(u_m)=1, \qquad \psi^{(m)}_{2n-1}(u_m)=\cos(n u_m), \qquad \psi^{(m)}_{2n}(u_m)=\sin(n u_m),
\label{eq:one_dimensional_blocks}
\end{equation}
for $n=1,\dots,N_f^{(m)}$. A separable feature is then uniquely identified by an integer multi-index $\mathbf{k} = (k_1, \dots, k_{N_\Pi})$:
\begin{equation}
\Psi_{\mathbf{k}}(u) = \prod_{m=1}^{N_\Pi}\psi^{(m)}_{k_m}(u_m), \qquad \mathbf{k}\in\mathcal{I}_{\mathbf{k}},
\label{eq:tensor_product_feature}
\end{equation}
where $\mathcal{I}_{\mathbf{k}}\subseteq\bigtimes_{m=1}^{N_\Pi}\{0,1,\dots,2N_f^{(m)}\}$ is the retained set of basis-function index vectors. The modulation is approximated as
\begin{equation}
\Phi(u)\approx \sum_{\mathbf{k}\in\mathcal{I}_{\mathbf{k}}} w_{\mathbf{k}} \Psi_{\mathbf{k}}(u),
\label{eq:phi_tensor_expansion}
\end{equation}
where the constant term corresponds to $\mathbf{k}=\mathbf{0}$, giving $w_{\mathbf{0}}\Psi_{\mathbf{0}}(u)=w_{\mathbf{0}}$.
This tensor-product representation allows different truncation orders in different coordinates and separates marginal effects from interaction terms. As shown later in the double-pendulum benchmark, Section~\ref{sec:doublependulum_results}, this distinction matters in multi-invariant settings with directional structure.

\subsection{Learning by regularized regression}
\label{sec:regularized_regression}

We define the feature vector $ \varphi(\mathbf{x}) \in \mathbb{R}^{p} $ as the concatenation of the basis functions for each of the $ K $ candidate monomial terms:
\begin{equation}
\varphi(\mathbf{x}) = \left[ \varphi_1(\mathbf{x})^\top, \varphi_2(\mathbf{x})^\top, \dots, \varphi_K(\mathbf{x})^\top \right]^\top,
\label{eq:feature_vector_structure}
\end{equation}
where each block $ \varphi_\ell(\mathbf{x}) $ contains the features associated with the $ \ell $-th monomial. Here, $P_\ell \equiv P_{\bm{\beta}_\ell}$ corresponds to the $\ell$-th admissible exponent vector $\bm{\beta} \in \mathcal{T}$ (where $\mathcal{T}\subset\mathcal{S}_b$) identified in the prefactor search. 

The construction of the feature elements is governed by the specific geometry of the selected dictionary. In the case of a phase-combination dictionary defined by multi-indices $\mathbf{n} \in \mathcal{I}_{\mathbf{n}}$, the feature set comprises the constant dimensional term $P_\ell(\mathbf{x})$ and the corresponding trigonometric pairs $P_\ell(\mathbf{x}) \cos(\mathbf{n} \cdot u)$ and $P_\ell(\mathbf{x}) \sin(\mathbf{n} \cdot u)$. Alternatively, when using a separable tensor-product dictionary, the features are indexed by $\mathbf{k} \in \mathcal{I}_{\mathbf{k}}$ and defined as $\varphi_{\ell,\mathbf{k}}(\mathbf{x}) = P_\ell(\mathbf{x}) \Psi_{\mathbf{k}}(u)$, where $\Psi_{\mathbf{k}}$ represents the product of the independent one-dimensional marginal basis functions associated with each invariant coordinate.

For a dataset of $N_{\mathrm{dat}}$ observations and a fixed candidate prefactor $\bm{\beta}$, we construct the design matrix $\mathcal{X}_{\bm{\beta}} \in \mathbb{R}^{N_{\mathrm{dat}} \times p}$ such that $[\mathcal{X}_{\bm{\beta}}]_{i, \cdot} = \varphi_{\bm{\beta}}(\mathbf{x}^{(i)})^\top$, and define the target vector $\mathbf{y} = [y^{(1)}, \dots, y^{(N_{\mathrm{dat}})}]^\top$. The predictor $ g(\mathbf{x}) = \mathbf{w}^\top \varphi(\mathbf{x}) $ is linear in the global coefficient vector $ \mathbf{w} \in \mathbb{R}^p $. Scalar weights for individual dictionary modes are written as follows: $w_n^{c}$, $w_n^{s}$ for cosine and sine modes in the one-dimensional case; $w_{\mathbf{n}}^{c}$, $w_{\mathbf{n}}^{s}$ for the phase-combination basis indexed by the frequency vector $\mathbf{n}$; and $w_{\mathbf{k}}$ for the separable tensor-product basis, where each index $\mathbf{k}$ already identifies a specific trigonometric factor through the basis functions $\psi^{(m)}_{k_m}$ defined in~\eqref{eq:one_dimensional_blocks}. The coefficients are estimated by minimizing the Ridge~\cite{hoerl1970} objective:
\begin{equation}
\hat{\mathbf{w}}_{\bm{\beta}} = \arg\min_{\mathbf{w} \in \mathbb{R}^p} \left[ \| \mathbf{y} - \mathcal{X}_{\bm{\beta}} \mathbf{w} \|_2^2 + \mu \|\mathbf{w}\|_2^2 \right].
\label{eq:ridge_objective}
\end{equation}
{The $\ell_2$ penalty damps high-frequency coefficients, and the objective remains quadratic in the weights. The implementation details and cross-validation protocol used in the numerical experiments are described in Section~\ref{sec:exp_protocol}.}

\subsubsection{Feature count and truncation}\label{sec:feature_count_and_truncation}
The dimensionality $p$ of the regression problem is a direct consequence of the truncation strategy employed in the invariant space. For a phase-combination dictionary with rectangular truncation and frequency limit $N_f$, the number of non-zero multi-indices is $|\mathcal{I}_{\mathbf{n}}|=(N_f+1)^{N_\Pi}-1$, resulting in $p=K(2(N_f+1)^{N_\Pi}-1)$ parameters. This exponential scaling with $ N_\Pi $ is the primary bottleneck for complex systems. To overcome this, we adopt a total-degree truncation $ \|\mathbf{k}\|_1 \le N_{\mathrm{tot}} $. The number of non-negative integer multi-indices satisfying this bound is given by the binomial coefficient $ \binom{N_{\mathrm{tot}} + N_\Pi}{N_\Pi} $. Including the constant term and the sine-cosine pairs, the total parameter count grows as:
\begin{equation}
p = K \left( 2 \binom{N_{\mathrm{tot}} + N_\Pi}{N_\Pi} - 1 \right) \approx \mathcal{O}\left(K \cdot \frac{N_{\mathrm{tot}}^{N_\Pi}}{N_\Pi!}\right).
\label{eq:polynomial_scaling}
\end{equation}
This specific combinatorics formula applies to the phase-combination basis under total-degree truncation. By contrast, for a separable tensor-product dictionary without symmetry reductions, the feature count scales as $ p = K \prod_{m=1}^{N_\Pi} (2N_f^{(m)} + 1) $. { This follows because each invariant coordinate contributes one constant feature plus $N_f^{(m)}$ cosine modes and $N_f^{(m)}$ sine modes, giving $2N_f^{(m)}+1$ one-dimensional basis functions per coordinate. The separable dictionary is obtained by taking all products of these one-dimensional factors, hence the product over $m$.} In both cases, the model is linear in the fitted weights.

\begin{figure}[h!]
\centering
\resizebox{0.8\textwidth}{!}{
\begin{tikzpicture}[
    font=\sffamily\small,
    % Main box styles
    dataNode/.style={rectangle, draw=green!60!black, fill=green!5, thick, rounded corners, align=center, minimum width=5.5cm, minimum height=1.2cm, inner sep=6pt},
    dynamicsNode/.style={rectangle, draw=blue!70!black, fill=blue!5, thick, rounded corners, align=left, minimum width=6.5cm, minimum height=1.2cm, inner sep=8pt},
    scaleNode/.style={rectangle, draw=orange!80!black, fill=orange!5, thick, rounded corners, align=left, minimum width=5cm, minimum height=1.2cm, inner sep=8pt},
    regressNode/.style={rectangle, draw=red!70!black, fill=red!5, thick, rounded corners, align=center, minimum width=8cm, minimum height=1.5cm, inner sep=8pt},
    % Arrow style
    myArrow/.style={-{Stealth[scale=1.2]}, thick, draw=black!70, rounded corners=4pt},
    % Annotation text on arrows
    arrowLabel/.style={fill=white, font=\itshape\scriptsize, text=black!80, inner sep=2pt, align=center},
    % Step numbers
    stepBadge/.style={circle, draw=white, fill=#1!80!black, text=white, font=\bfseries\footnotesize, inner sep=1pt, minimum size=8mm}
]

% ==========================================
% 1. TOP: INPUT DATA
% ==========================================
\node[dataNode] (input) {
    \textbf{1. Raw Dimensional Data}\\
    \vspace{2pt}
    \begin{tabular}{ll}
        \textit{Inputs:} & $\mathbf{x} \in \mathbb{R}^N$ \\
        \textit{Target:} & $y \in \mathbb{R}$ \\
        \textit{Matrix:} & $A \in \mathbb{Q}^{R \times N}$
    \end{tabular}
};
\node[stepBadge=green, xshift=-1cm] at (input.west) {1};

% ==========================================
% 2. THE SPLIT
% ==========================================

% --- LEFT TRACK: DYNAMICS (pi groups) ---
\node[dynamicsNode, below left=1.8cm and 0.5cm of input] (kernel) {
    \textbf{2a. Buckingham Invariants}\\
    \vspace{2pt}
    \footnotesize
    Compute basis for $\ker(A)$:\\
    $\pi_m = \prod_j x_j^{\gamma_{j}^{(m)}} \quad (\text{where } A\bm{\gamma} = \bm{0})$
};
\node[stepBadge=blue, left=0pt of kernel.west, xshift=-3mm] {2a};

\node[dynamicsNode, below=0.8cm of kernel] (embed) {
    \textbf{2b. Affine Embedding}\\
    \vspace{2pt}
    \footnotesize
    Map to bounded domain $N_\Pi$-Torus:\\
    $u_m(\pi_m)
=
2\pi\,
\left(\frac{\pi_m-\pi_m^{\min}}{\pi_m^{\max}-\pi_m^{\min}}\right) \in [0, 2\pi]$
};
\node[stepBadge=blue, left=0pt of embed.west, xshift=-3mm] {2b};

% --- The Dictionary Fork ---
\node[dynamicsNode, below=0.8cm of embed] (harmonic) {
    \textbf{2c. Harmonic Expansion Dictionary}\\
    \vspace{4pt}
    \footnotesize
    \textit{Option A: Separable Tensor-Product}\\
    $\Psi_{\mathbf{k}}(u) = \bigotimes_{m=1}^{N_{\Pi}}\begin{Bmatrix} \cos(k_m u_m) \\ \sin(k_m u_m) \end{Bmatrix}$\\
    \vspace{4pt}
    \footnotesize
    \textit{Option B: Phase-Combination}\\
    $\Psi_{\mathbf{n}}(u) \in \begin{Bmatrix} \cos(\mathbf{n} \cdot u) \\ \sin(\mathbf{n} \cdot u) \end{Bmatrix}$
};
\node[stepBadge=blue, left=0pt of harmonic.west, xshift=-3mm] {2c};

% --- RIGHT TRACK: SCALING (Prefactors) ---
\node[scaleNode, below right=1.8cm and 0.5cm of input] (prefactor) {
    \textbf{3. Dimensional Skeleton}\\
    \vspace{2pt}
    \footnotesize
    Solve affine system $A\bm{\beta} = \mathbf{b}$:\\
    $P_{\bm{\beta}}(\mathbf{x}) = \prod_{j=1}^N x_j^{\beta_j}$
};
\node[stepBadge=orange, right=0pt of prefactor.east, xshift=3mm] {3};

% ==========================================
% 3. BOTTOM: REGRESSION
% ==========================================
\node[regressNode, below right=1.8cm and 0.5cm of harmonic] (regression) {
    \textbf{4. Regularized RidgeCV Regression}\\
    \vspace{4pt}
    \small
    $ \hat{\mathbf{w}}_{\bm{\beta}} = \operatorname*{argmin}_{\mathbf{w}} \left\| \mathbf{y} - \mathcal{X}_{\bm{\beta}} \mathbf{w} \right\|_2^2 + \mu \|\mathbf{w}\|_2^2 $
};
\node[stepBadge=red, xshift=-1cm] at (regression.west) {4};

% ==========================================
% 4. ROUTING AND ARROWS
% ==========================================

% Split from Input
\draw[myArrow] (input.south) -- ++(0,-0.3) -| (kernel.north) 
    node[pos=0.25, arrowLabel, anchor=east, xshift=-4mm] {Isolate structural shape (dimensionless)};
    
\draw[myArrow] (input.south) -- ++(0,-0.3) -| (prefactor.north) 
    node[pos=0.25, arrowLabel, anchor=west, xshift=4mm] {Isolate physical units (dimensional)};

% Left track flow
\draw[myArrow] (kernel) -- (embed);
\draw[myArrow] (embed) -- (harmonic);

% Converge to Output
\draw[myArrow] (harmonic.south) -- ++(0,-1.3) -| (regression.north)
    node[pos=0.25, arrowLabel, anchor=east, xshift=-4mm] {Matrix of harmonic features\\$\Psi(u)$};

% Calculate proper drop for the right side
\draw[myArrow] (prefactor.south) -- ++(0,-5.2) -| (regression.north)
    node[pos=0.25, arrowLabel, anchor=west, xshift=4mm] {Static scaling vector\\$P_{\bm{\beta}}(\mathbf{x})$};

% ==========================================
% 5. BACKGROUND ZONES
% ==========================================
\begin{scope}[on background layer]
    % Left Background (Dynamics)
    \node[fill=blue!3, rounded corners=8pt, draw=blue!20, dashed, 
          fit=(kernel) (embed) (harmonic) (kernel.west|-kernel.north), 
          inner xsep=18pt, inner ysep=14pt] (leftZone) {};
    \node[anchor=south west, font=\bfseries\scriptsize, text=blue!80!black] 
          at (leftZone.north west) {PHASE 2: Dimensionless Modulations};
          
    % Right Background (Scale)
    \node[fill=orange!3, rounded corners=8pt, draw=orange!20, dashed, 
          fit=(prefactor) (prefactor.east|-prefactor.north), 
          inner xsep=18pt, inner ysep=14pt] (rightZone) {};
    \node[anchor=south east, font=\bfseries\scriptsize, text=orange!80!black] 
          at (rightZone.north east) {PHASE 3: Dimensional Units};
\end{scope}

\end{tikzpicture}
}
\caption{Algorithmic pipeline. Box 2c illustrates the two harmonic dictionary formulations evaluated in the experiments: separable tensor-product (option $A$) and phase-combination (option $B$) dictionaries.}
\label{fig:algorithmic_pipeline}
\end{figure}
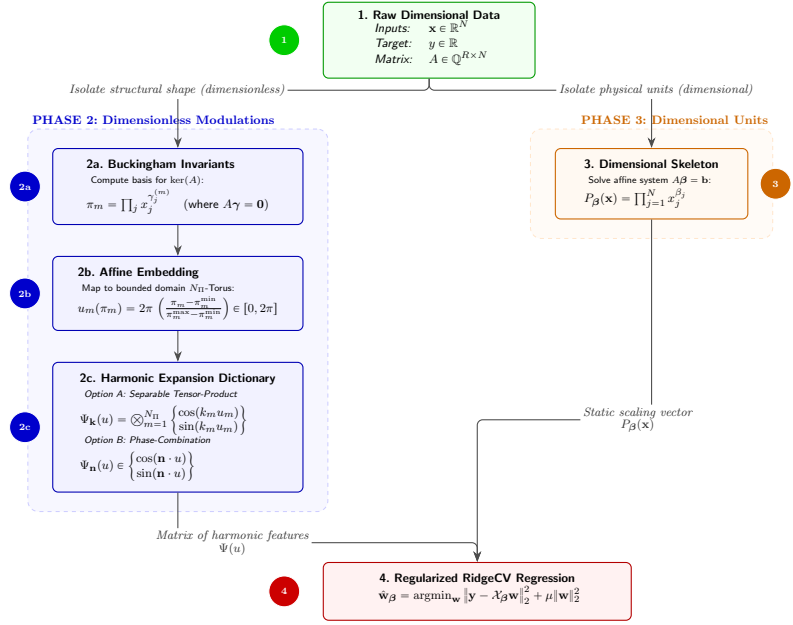

\newpage

\section{Algorithmic pipeline}
\label{sec:algorithm}

{We now summarize the computational pipeline that turns raw measurements into
an explicit, dimensionally consistent predictor. The purpose of this section is
not to repeat the mathematical construction of
Section~\ref{sec:learning_physical_laws}, but to make explicit the operational
workflow and the computational scaling of the different stages of the method.

Concretely, the algorithm proceeds by (i) constructing independent $\Pi$ groups,
(ii) building dimensionally admissible prefactors with the dimensions of $y$,
(iii) expanding the unknown dimensionless modulation in a truncated dictionary
on invariant space, and (iv) fitting the resulting linear-in-parameters model
by regularized regression.}

\begin{algorithm}[H]
  \caption{Unit-equivariant learning via tensor-product harmonic dictionaries}
  \label{alg:full_pipeline}
\resizebox{0.95\textwidth}{!}{%
\begin{minipage}{1.05\textwidth}
\begin{algorithmic}[1]
  \Require Dataset $\mathcal{D} = \{(\mathbf{x}^{(i)}, y^{(i)})\}_{i=1}^{N_{\mathrm{dat}}}$; Truncation orders $\{N_f^{(m)}\}_{m=1}^{N_\Pi}$ (for separable basis, index set $\mathcal{I}_{\mathbf{k}}$; for phase-combination basis, index set $\mathcal{I}_{\mathbf{n}}$); Regularization grid $\Lambda$.
  \Ensure Fitted analytic predictor $\hat{g}(\mathbf{x})$.

  \Statex \textbf{Phase 1: Dimensionless Manifold Mapping}
  \State Compute basis $\{\bm{\gamma}^{(m)}\}_{m=1}^{N_\Pi}$ for $\ker(A)$.
  \State Construct independent invariants: $\pi_m(\mathbf{x}) = \prod_{j=1}^N x_j^{\gamma_j^{(m)}}$ for $m = 1, \dots, N_\Pi$.
  \State Compute affine embedding to map $\pi(\mathbf{x}^{(i)})$ to periodic coordinates $u^{(i)} \in [0, 2\pi]^{N_\Pi}$.

  \Statex \textbf{Phase 2: Tensor-Product Dictionary Construction}
  \State Initialize dictionary $\mathcal{I}_{\mathbf{k}}$ based on symmetry reductions
       (e.g., retaining only even-parity tensor-product features).
  \For{each coordinate $m \in \{1, \dots, N_\Pi\}$}
      \State Generate 1D marginal basis: $\mathcal{F}_m = \{1\} \cup \{\cos(n u_m), \sin(n u_m)\}_{n=1}^{N_f^{(m)}}$.
  \EndFor
  \State Construct separable basis $\Psi_{\mathbf{k}}(u)$ via tensor product: $\bigotimes_{m=1}^{N_\Pi} \mathcal{F}_m$.
  \State (Optional) Apply total-degree truncation (see section \ref{sec:feature_count_and_truncation}): restrict to $\|\mathbf{k}\|_1 \le N_{\mathrm{tot}}$.

  \Statex \textbf{Phase 3: Structural Optimization \& Regression}
  \State Identify affine subspace of admissible exponents: $\mathcal{S}_b = \{\bm{\beta} \in \mathbb{Q}^N \mid A\bm{\beta} = \mathbf{b}\}$.
  \State Generate search grid $\mathcal{T} \subset \mathcal{S}_b$ for candidate dimensional prefactors.
  \For{each candidate exponent vector $\bm{\beta} \in \mathcal{T}$}
      \State Construct dimensional skeleton $P_{\bm{\beta}}(\mathbf{x})$.
      \State Build design matrix $\mathcal{X}_{\bm{\beta}}$ with elements $\left[\mathcal{X}_{\bm{\beta}}\right]_{i, \mathbf{k}} = P_{\bm{\beta}}(\mathbf{x}^{(i)}) \Psi_{\mathbf{k}}(u^{(i)})$ for $\mathbf{k} \in \mathcal{I}_{\mathbf{k}}$.
      \State Solve regularized regression via cross-validation over $\Lambda$:
             $$ \hat{\mathbf{w}}_{\bm{\beta}} = \operatorname*{argmin}_{\mathbf{w}} \left\| \mathbf{y} - \mathcal{X}_{\bm{\beta}} \mathbf{w} \right\|_2^2 + \mu \|\mathbf{w}\|_2^2 $$
      \State Record cross-validation metric $\mathcal{M}(\bm{\beta})$ (e.g., $R^2$ or negative MSE).
  \EndFor
  \State Select optimal structural skeleton: $\bm{\beta}^* = \operatorname*{argmax}_{\bm{\beta} \in \mathcal{T}} \mathcal{M}(\bm{\beta})$.

  \Statex \textbf{Phase 4: Symbolic Recovery}
  \State \Return $\hat{g}(\mathbf{x}) = P_{\bm{\beta}^*}(\mathbf{x}) \sum_{\mathbf{k} \in \mathcal{I}_{\mathbf{k}}} \hat{w}_{\bm{\beta}^*, \mathbf{k}} \Psi_{\mathbf{k}}(u(\mathbf{x}))$.
\end{algorithmic}
\end{minipage}
}
\end{algorithm}
\noindent\textit{Remark.}
If a phase-combination basis is used instead of the separable tensor-product
dictionary, the dictionary-construction step is replaced by the direct
evaluation of the features
$w_{\mathbf{n}}^{c}\cos(\mathbf{n}\cdot u)$ and
$w_{\mathbf{n}}^{s}\sin(\mathbf{n}\cdot u)$ for
$\mathbf{n}\in\mathcal{I}_{\mathbf{n}}$, where
$\mathbf{n}\cdot u=\sum_m n_m u_m$.

\subsection{Computational complexity and scalability}
The prefactor-selection step requires evaluating the Ridge regression with leave-one-out cross-validation (\texttt{RidgeCV}) fitting procedure
for each point on a grid over the affine subspace $\mathcal{S}_b$. For a system
with $N_\Pi$ free exponents in the prefactor family, the grid has
$\mathcal{O}(M^{N_\Pi})$ evaluations, where $M$ is the number of grid points per
dimension. Each evaluation involves a \texttt{RidgeCV} fit with cost
$\mathcal{O}(N_{\mathrm{dat}}p^2+p^3)$, where $p$ is the number of dictionary
features. The total cost therefore scales as $\mathcal{O}\!\left(
M^{N_\Pi}\left(N_{\mathrm{dat}}p^2+p^3\right)
\right)$. The cost $\mathcal{O}(N_{\mathrm{dat}}p^2+p^3)$ arises from the standard
regularized least-squares solution. Forming the covariance matrix
$\mathcal{X}^{\top}\mathcal{X}$ requires
$\mathcal{O}(N_{\mathrm{dat}}p^2)$ operations, and the subsequent inversion or
decomposition of this $p\times p$ matrix scales as $\mathcal{O}(p^3)$. Since
\texttt{RidgeCV} employs an efficient leave-one-out cross-validation scheme
that reuses the same matrix decomposition for multiple values of the
regularization parameter $\mu$, the overall complexity per grid point remains
dominated by these two terms \cite{golub2013}. {For the benchmarks considered in this work ($N_\Pi\leq 2$), this exhaustive
scan is computationally feasible. In practical terms, the current
implementation is intended for low-dimensional invariant spaces, roughly
$N_\Pi\leq 3$, with moderate truncation orders and prefactor grids. For
$N_\Pi\geq 4$, the combination of the prefactor grid and high-dimensional
dictionaries becomes the main computational bottleneck, since both the number
of candidate prefactors and the number of fitted features can grow rapidly.

Several strategies can mitigate this limitation. On the prefactor side, the
brute-force scan over $\mathcal{S}_b$ could be replaced by gradient-based
optimization over the exponent parameters~\citep{bengio2000gradient},
coarse-to-fine hierarchical grid searches~\citep{scikit-learn}, or
Bayesian optimization over the prefactor space~\citep{frazier2018tutorial}. On
the dictionary side, sparse total-degree truncations, adaptive frequency
selection, or regularization paths could be used to avoid fitting the full
dictionary. These extensions are outside the scope of the present benchmarks,
but they provide natural routes for scaling the method to higher-dimensional
invariant spaces.}

\section{Experimental setup}
\label{sec:experiments}

{We evaluate the method on three synthetic benchmarks and one experimental black-body spectrum dataset.} The experiments are designed to examine dimensional constraints, dictionary choice, and prefactor selection in different regimes: a case in which dimensional analysis nearly determines the response, a case with a strongly non-polynomial modulation, and a case with a structured multi-invariant response surface.

\newpage

\subsection{General protocol}
\label{sec:exp_protocol}

For each benchmark, we construct a dataset of input-output pairs $\mathcal{D}=\{(\mathbf{x}^{(i)},y^{(i)})\}_{i=1}^{N_{\mathrm{dat}}}$, and divide it into training and test subsets in order to evaluate out-of-sample performance. When the dimensional prefactor is not unique, prefactor selection is treated as an outer model-selection loop: each candidate prefactor defines a different feature map, the corresponding harmonic coefficients are fitted by regularized regression, and the final model is chosen according to validation performance.

Predictive performance is quantified primarily through the coefficient of determination $R^2$ and the mean squared error (MSE),
\begin{align}
R^2 = 1-\frac{\sum_{i=1}^{N_{\mathrm{test}}}(y^{(i)}-\hat{y}^{(i)})^2}{\sum_{i=1}^{N_{\mathrm{test}}}(y^{(i)}-\bar{y})^2}, \qquad \mathrm{MSE} = \frac{1}{N_{\mathrm{test}}}\sum_{i=1}^{N_{\mathrm{test}}}(y^{(i)}-\hat{y}^{(i)})^2,
\end{align}
where $y^{(i)}$ denotes the true target, $\hat{y}^{(i)}$ the prediction, $\bar{y}$ the empirical mean of the test targets, and $N_{\mathrm{test}}$ the size of the evaluation set.

Unless otherwise stated, the fitting stage is carried out using regularized linear models implemented in \texttt{scikit-learn}~\cite{scikit-learn}. Throughout all experiments, \texttt{RidgeCV} is used by default. The regularization parameter $\mu$ is selected by cross-validation within the training set only. Across all experiments, we study the framework under variations in four factors:
\begin{itemize}
    \item the dimensional complexity of the benchmark;
    \item the number of retained invariant coordinates;
    \item the richness of the harmonic dictionary;
    \item the sampling density and the level of additive noise.
\end{itemize}
This makes it possible to assess separately the effects of dimensional structure, approximation bias, and statistical variability.

\subsection{Benchmarks}
\label{sec:benchmark_overview}

The benchmarks considered in this work are: (i) a small-angle simple pendulum,
used to isolate recovery of the dimensional scaling; (ii) synthetic
black-body radiation at fixed frequency, used to test a one-dimensional but
transcendental residual modulation; (iii) the COBE/FIRAS black-body spectrum,
used as a real-data validation and as a comparison between harmonic and
Chebyshev features; and (iv) the double-pendulum Lyapunov field, used to test
dictionary geometry, tensor-product structure and symmetry reduction in a
multi-invariant setting.

Taken together, the experiments are designed to answer four questions: whether
the framework recovers the correct dimensional scaling, how prefactor selection
interacts with approximation error, how dictionary geometry affects
multi-invariant learning, and how regularization and symmetry reduction improve
robustness.

\section{Results}
\label{sec:results}

In this section, we report the results for each benchmark.

\subsection{Simple pendulum}
\label{sec:pendulum_unified}

{
We begin with the simple pendulum as a dimensional-skeleton benchmark. With the
release angle fixed, the learning task reduces to recovering the scaling
$\tau\,\propto\,\sqrt{L/g}$ and testing its robustness under different sampling
and noise regimes.}

\subsubsection{Physical setting and dimensional structure}

We consider a simple pendulum consisting of a point mass $m$ suspended by a rigid massless rod of length $L$ in a uniform gravitational field $g$. For release angle $\theta_0$, the period is described by
{
\begin{align}
\tau
=
2\pi\sqrt{\frac{L}{g}}\,
\frac{2}{\pi}
\mathcal{K}\!\left(\sin\frac{\theta_0}{2}\right),
\end{align}}
where $\mathcal{K}$ is the complete elliptic integral of the first kind 
{$$\mathcal{K}(x) = \int_{0}^{\pi/2} \frac{d\theta}{\sqrt{1 - x \sin^{2}\theta}}.$$}
In the small-angle regime $\theta_0\ll 1$, this reduces to the familiar approximation $\tau\approx 2\pi\sqrt{\frac{L}{g}}$. We model the period $\tau$ as a function of the variables $(L,g,m,\theta_0)$. 
The associated dimension matrix (if we let $\mathcal{B}=\{\mathsf{M},\mathsf{L},\mathsf{T}\}$) is
\begin{align}
A=
\begin{pmatrix}
0 & 0 & 1 & 0 \\
1 & 1 & 0 & 0 \\
0 & -2 & 0 & 0
\end{pmatrix}.
\end{align}
Since the angle is already dimensionless and the mass does not affect the period law, the invariant structure is particularly simple. Solving $A\bm{\gamma}=\mathbf{0}$ yields a one-dimensional invariant space, and a natural choice is $\pi_1=\theta_0$. Likewise, solving $A\bm{\beta}=\mathbf{b}$ for the target signature of $\tau$ gives the dimensional prefactor $P=\sqrt{\frac{L}{g}}$, up to multiplication by powers of the invariant, which can be absorbed into the dimensionless modulation. The resulting model therefore takes the form
\begin{equation}
\tau=\,\Phi(\theta_0)\,\sqrt{\frac{L}{g}}.
\label{eq:pendulum_factorization}
\end{equation}

\subsubsection{Model and data generation}

For this benchmark, the predictor is written as
\begin{equation}
\hat{\tau}
=
\sqrt{\frac{L}{g}}
\left[
w_0+\sum_{n=1}^{N_f}
\bigl(
w_n^{c}\cos(nu_1)+w_n^{s}\sin(nu_1)
\bigr)
\right],
\label{eq:pendulum_fourier_model}
\end{equation}
where $u_1$ is the affine rescaling of $\theta_0$ to the interval $[0,2\pi]$. For these experiments, the period of the pendulum is assumed to be $\tau = 2\pi\sqrt{\frac{L}{g}}$. { In this scenario, the dimensionless modulation is constant, $\Phi(\theta_0)=2\pi$. The experiment therefore mainly tests the effect of using the dimensional prefactor; the harmonic part of the model is exercised in the black-body and double-pendulum benchmarks, where the modulation is non-trivial.}

Synthetic data are generated from the exact pendulum law under a two-level sampling procedure. First, $n_L$ distinct pendulum lengths are drawn uniformly from the interval $[0.1,2.0]$ m. For each sampled length, $M$ repeated measurements are produced by adding Gaussian noise $\mathcal{N}(0,\sigma_{\mathrm{abs}}^2)$ to the ground-truth period, where $\sigma_{\mathrm{abs}}$ is an absolute noise level in seconds. The repeated measurements are then averaged before regression. This setup allows us to vary independently the measurement noise, the number of distinct length values, and the number of repeated observations.

To examine the stability of the framework, we consider six representative scenarios combining low and high noise with sparse and dense sampling. As a baseline comparison, we also report fits obtained with a standard \texttt{Minuit}~\cite{James:1975dr} parametric estimator.

\subsubsection{Results}

The numerical results are summarized in Table~\ref{tab:pendulum_results}. In low-noise regimes, both approaches recover the expected scaling with high accuracy. In particular, when the sampling density is sufficiently large, the learned coefficient is essentially indistinguishable from the theoretical small-angle value $2\pi\approx 6.2832$.

For comparison with the theoretical small-angle scaling, we report the coefficient ($c$) and intercept ($d$) obtained from an auxiliary post hoc fit of the form $\tau = c\,\sqrt{L/g}+d$.

\begin{table}[htbp]
\centering
\caption{Simple-pendulum results comparing regularized regression and \texttt{Minuit}. The theoretical small-angle coefficient is $2\pi \approx 6.2832$.}
\label{tab:pendulum_results}

\footnotesize
\setlength{\tabcolsep}{4pt}
\renewcommand{\arraystretch}{1.08}

\begin{tabular}{l c c l c c c}
\toprule
Scenario & $\sigma_{\mathrm{abs}}$ & $(n_L,M)$ & Method & Coeff.\ $(c)$ & Intercept $(d)$ & Error (\%) \\
\midrule

\multirow{2}{*}{Low Noise}
& \multirow{2}{*}{0.02}
& \multirow{2}{*}{(5,5)}
& \texttt{RidgeCV} & 6.2539 & 0.0061 & 0.47 \\
& & & \texttt{Minuit}  & $6.2451 \pm 0.0287$ & $0.0079 \pm 0.0011$ & 0.61 \\
\addlinespace

\multirow{2}{*}{High Density}
& \multirow{2}{*}{0.02}
& \multirow{2}{*}{(50,50)}
& \texttt{RidgeCV} & 6.2836 & 0.0007 & 0.01 \\
& & & \texttt{Minuit}  & $6.2836 \pm 0.0142$ & $0.0006 \pm 0.0053$ & 0.01 \\
\addlinespace

\multirow{2}{*}{High Noise}
& \multirow{2}{*}{0.6}
& \multirow{2}{*}{(5,5)}
& \texttt{RidgeCV} & 5.4052 & 0.1822 & 13.97 \\
& & & \texttt{Minuit}  & $5.1418 \pm 0.8622$ & $0.2382 \pm 0.3164$ & 18.16 \\
\addlinespace

\multirow{2}{*}{High Density Noise}
& \multirow{2}{*}{0.6}
& \multirow{2}{*}{(50,50)}
& \texttt{RidgeCV} & 6.29028 & 0.01799 & 0.10 \\
& & & \texttt{Minuit}  & $6.2916 \pm 0.4266$ & $0.0182 \pm 0.1588$ & 0.20 \\
\addlinespace

\multirow{2}{*}{Mixed (Low $n_L$)}
& \multirow{2}{*}{0.6}
& \multirow{2}{*}{(5,50)}
& \texttt{RidgeCV} & 6.5910 & -0.0997 & 4.90 \\
& & & \texttt{Minuit}  & $6.5941 \pm 1.1526$ & $-0.0981 \pm 0.4282$ & 4.95 \\
\addlinespace

\multirow{2}{*}{Mixed (Low $M$)}
& \multirow{2}{*}{0.6}
& \multirow{2}{*}{(50,5)}
& \texttt{RidgeCV} & 6.55672 & -0.08879 & 4.52 \\
& & & \texttt{Minuit}  & $6.5411 \pm 0.3375$ & $-0.1078 \pm 0.1217$ & 4.11 \\
\bottomrule
\end{tabular}
\end{table}

The main qualitative trend is that the dimensionally constrained regression model remains accurate whenever either the sampling density or the amount of averaging is sufficient. Under sparse and noisy conditions, both methods deteriorate, but the regularized estimator is generally more stable. This is particularly visible in the high-noise, low-sampling setting, where Ridge regression yields a smaller error than the baseline parametric fit. The effect is consistent with the variance-reduction role of the $\ell_2$ penalty.

\begin{figure}[htbp]
  \centering
  \captionsetup[subfigure]{width=0.8\linewidth}
  % --- Row 1 ---
  \begin{subfigure}[t]{0.3333\textwidth}
    \centering
    \includegraphics[width=\textwidth]{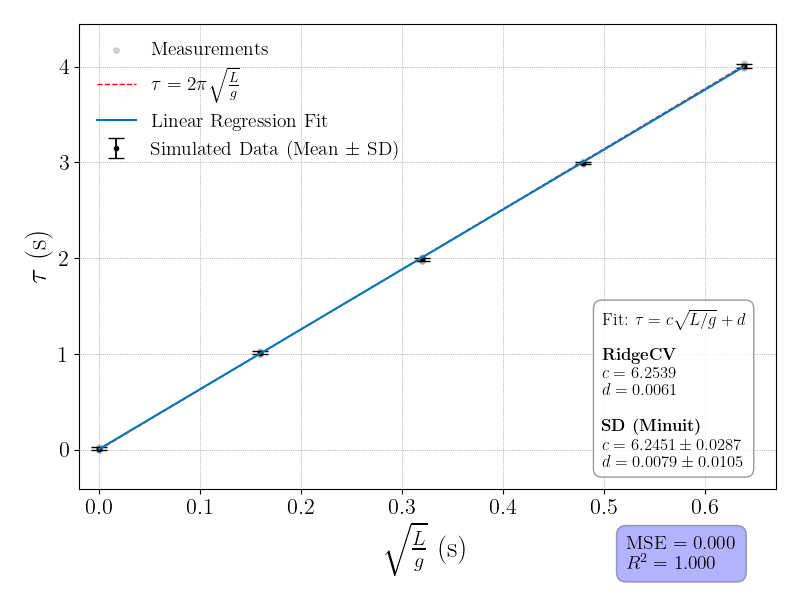}
    \caption{Low noise; $\sigma_{\text{abs}}=0.02$, $n_L=M=5$.}
    \label{fig:low_noise_low_data}
  \end{subfigure}\hfill
  \begin{subfigure}[t]{0.3333\textwidth}
    \centering
    \includegraphics[width=\textwidth]{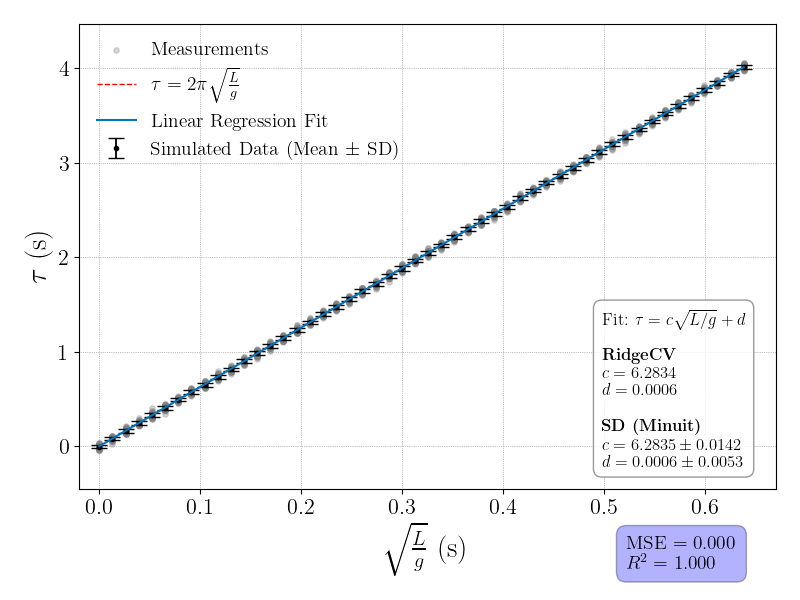}
    \caption{High density; $\sigma_{\text{abs}}=0.02$, $n_L=M=50$.}
    \label{fig:low_noise_high_data}
  \end{subfigure}\hfill
  \begin{subfigure}[t]{0.3333\textwidth}
    \centering
    \includegraphics[width=\textwidth]{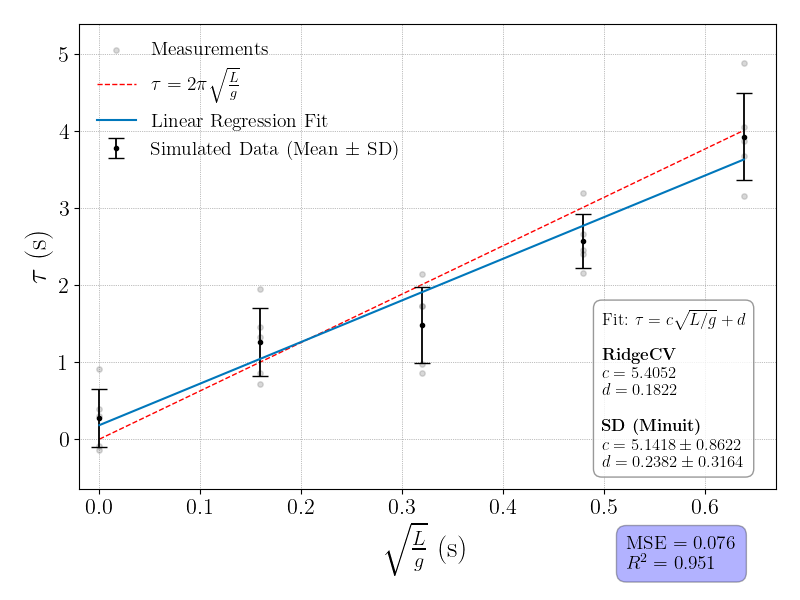}
    \caption{High noise; $\sigma_{\text{abs}}=0.6$, $n_L=M=5$.}
    \label{fig:high_noise_low_data}
  \end{subfigure}

  \vspace{1.5em} % More space to prevent captions from hitting the next row

  % --- Row 2 ---
  \begin{subfigure}[t]{0.3333\textwidth}
    \centering
    \includegraphics[width=\textwidth]{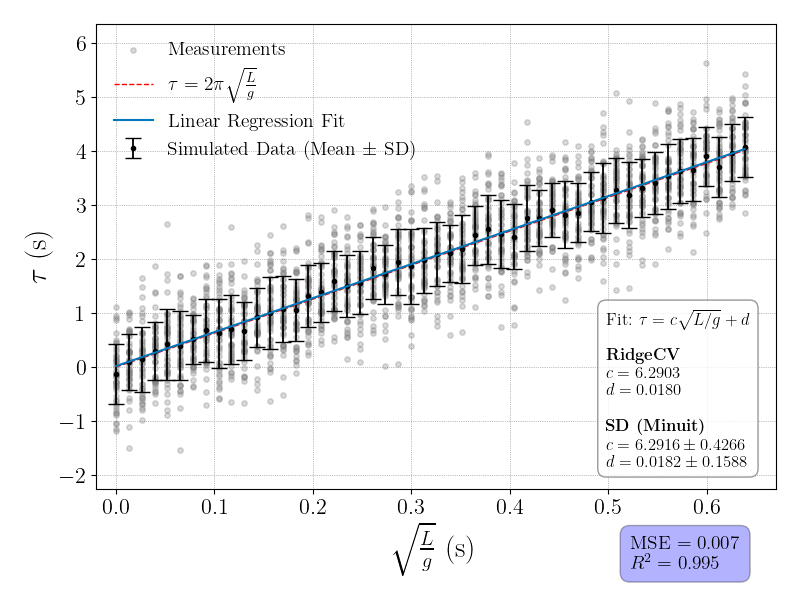}
    \caption{High density noise; $\sigma_{\text{abs}}=0.6$, $n_L=M=50$.}
    \label{fig:high_noise_high_data}
  \end{subfigure}\hfill
  \begin{subfigure}[t]{0.3333\textwidth}
    \centering
    \includegraphics[width=\textwidth]{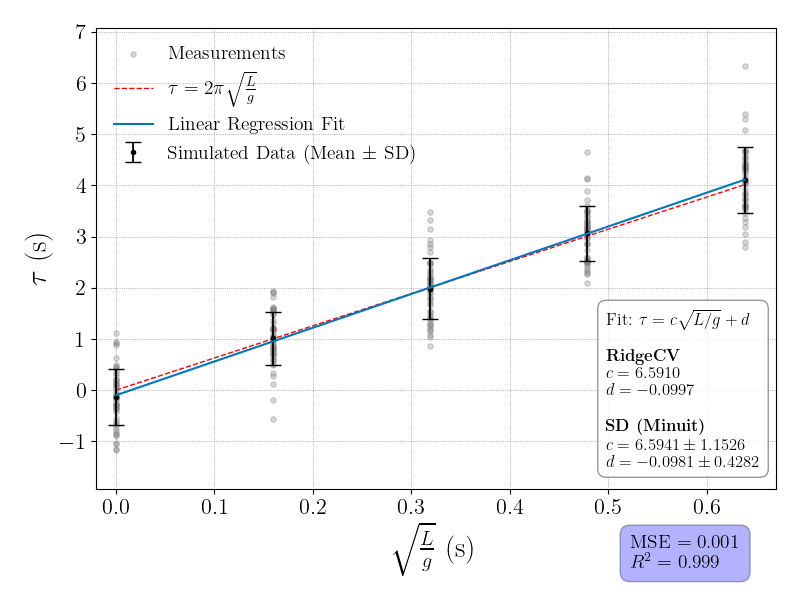}
    \caption{Mixed (low $n_L$); $\sigma_{\text{abs}}=0.6$, $n_L=5$, $M=50$.}
    \label{fig:high_noise_low_nl}
  \end{subfigure}\hfill
  \begin{subfigure}[t]{0.3333\textwidth}
    \centering
    \includegraphics[width=\textwidth]{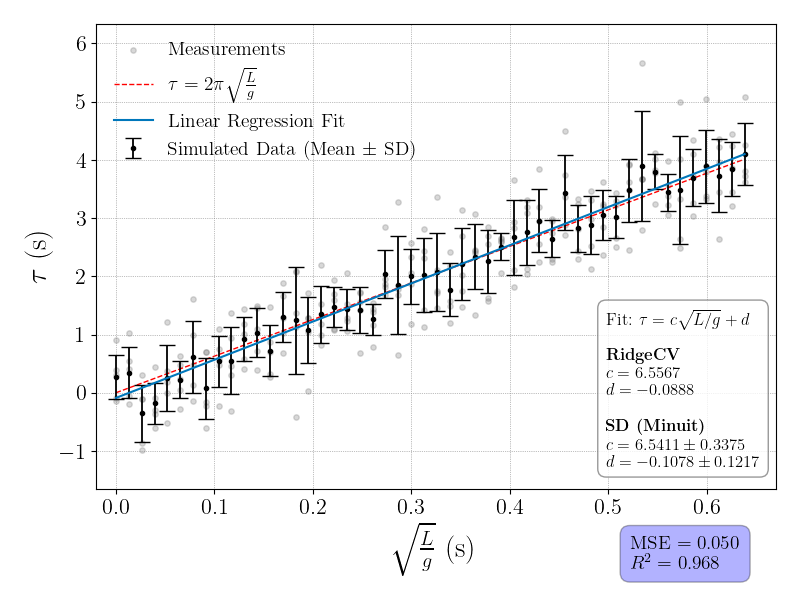}
    \caption{Mixed (low $M$); $\sigma_{\text{abs}}=0.6$, $n_L=50$, $M=5$.}
    \label{fig:high_noise_low_m}
  \end{subfigure}

  \caption{Regression results for the simple pendulum period under different noise and sampling regimes. The red dashed line indicates the theoretical small-angle coefficient $2\pi$.}
  \label{fig:pendulum_scenarios_combined}
\end{figure}

These trends are illustrated in Figure~\ref{fig:pendulum_scenarios_combined}. Increasing either the number of sampled lengths or the number of repeated measurements is enough to recover the correct scaling even when the noise level is large.

\subsubsection{Interpretation}

This benchmark behaves as expected in the simplest setting. The dimensional prefactor recovers the scaling $\sqrt{L/g}$, and the residual dimensionless dependence is constant in the small-angle approximation. The simple pendulum therefore serves mainly as a baseline check: when dimensional analysis almost determines the response, the model reproduces the expected scaling with little data.

{
\subsubsection{Comparison with an unconstrained baseline}

To quantify the benefit of imposing dimensional consistency, we compare the proposed model against an unconstrained power-law regressor with the same functional form. The dimensional model fixes the exponent from physics:
\begin{equation}
\hat\tau = w_0^{\rm (Dim)} + w_1^{\rm (Dim)}\,\sqrt{\frac{L}{g}},
\label{eq:dim_powerlaw}
\end{equation}
with 2 free parameters $(w_0,w_1)$. The unconstrained analog would be
$$\hat\tau = w_0^{\rm (Unc)} + w_1^{\rm (Unc)}\,L^{\alpha}\,g^\beta,
$$
Since $g$ is constant, we absorb it into $w_1$. So the final unconstrained polynomial is of the form:
\begin{equation}
\hat\tau = w_0^{\rm (Unc)} + w_1^{\rm (Unc)}\,L^{\alpha},
\label{eq:unc_powerlaw}
\end{equation}
with 3 parameters $(w_0,w_1,\alpha)$. Both models are fitted entirely via \texttt{Minuit}~\cite{James:1975dr}. For the dimensional model, Minuit directly minimizes the mean squared error (MSE) over $(w_0,w_1)$ on the training set. For the unconstrained model, we use a nested 80/20 train--validation split of the training data: \texttt{Minuit} optimizes all three parameters $(\alpha,w_0,w_1)$ simultaneously against the validation MSE. After convergence, the final out-of-sample $R^2$ is evaluated on the held-out test set using the discovered parameters. Both models use identical starting values $(w_0,w_1)=(0,0)$, and for the unconstrained model $\alpha$ is initialized at 0 and constrained to $[-2,2]$, with no prior knowledge of the physical exponent. All experiments use 30 independent 70/30 train/test trials.

In this comparison, fixing $\alpha=1/2$ removes the nonlinear optimization over the exponent and leaves a two-parameter linear fit. This allows the dimensional model to work from as few as $n_L=5$ data points. The unconstrained model requires a validation split to guide Minuit over $\alpha$ and fails for $n_L<12$, where the split leaves too few points to form reliable estimates.

As shown in Table~\ref{tab:unconstrained_comparison}, the unconstrained model requires substantial data before the exponent converges: at $n_L=12$, $\hat\alpha=0.796\pm0.140$ is far from the physical value and the out-of-sample $R^2$ is only $0.31$, reflecting the difficulty of jointly resolving three parameters from a validation set of just two points. With increasing $n_L$, $\hat\alpha$ drifts steadily toward $1/2$, reaching $0.556\pm0.099$ at $n_L=35$ and $0.526\pm0.032$ at $n_L=100$, while the $R^2$ rises to $0.999$. The dimensional model, by contrast, reaches $R^2\approx 0.999$ from $n_L=5$ onward. These results confirm that the dimensional constraint is \emph{sufficient but not strictly necessary} for this benchmark, a blind 3-parameter search eventually converges to the correct exponent, but requires an order of magnitude more data and passes through a regime of poor generalization where the joint optimization over $(\alpha,w_0,w_1)$ is ill-conditioned. The dimensional model's advantage is therefore one of sample efficiency, robustness and correctness.

\begin{table}[htbp]
\centering
\caption{Dimensional versus unconstrained power-law regression for the simple pendulum. Mean out-of-sample $R^2$ and recovered exponent $\hat\alpha$ over 30 independent trials. Both models fitted with \texttt{Minuit}.}
\label{tab:unconstrained_comparison}
\footnotesize
\setlength{\tabcolsep}{12pt}
\renewcommand{\arraystretch}{1.08}
\begin{tabular}{c c c c}
\toprule
$n_L$ & $R^2$ (Dim.) & $R^2$ (Unc.) & $\hat\alpha \pm \sigma_\alpha$ \\
\midrule
5   & 0.9982 & --- & --- \\
8   & 0.9977 & --- & --- \\
12  & 0.9985 & 0.3061 & $0.796 \pm 0.140$ \\
20  & 0.9991 & 0.9667 & $0.678 \pm 0.132$ \\
35  & 0.9995 & 0.9930 & $0.556 \pm 0.099$ \\
60  & 0.9994 & 0.9983 & $0.528 \pm 0.034$ \\
100 & 0.9995 & 0.9988 & $0.526 \pm 0.032$ \\
\bottomrule
\end{tabular}
\end{table}

The trends are illustrated in Figure~\ref{fig:unconstrained_baseline}. The dimensional model reaches $R^2\approx 0.999$ with as few as $n_L=5$ lengths, whereas the unconstrained model requires $n_L\geq 12$ to produce any result and $n_L\geq 60$ to approach the dimensional accuracy level. At small $n_L$, the simultaneous optimization over $(\alpha,w_0,w_1)$ is severely ill-conditioned, producing $\hat\alpha$ far from $1/2$ and low out-of-sample $R^2$; the convergence toward the physical exponent is gradual, requiring an order of magnitude more data than the dimensional model.

\begin{figure}[htbp]
  \centering
  \includegraphics[width=1\textwidth]{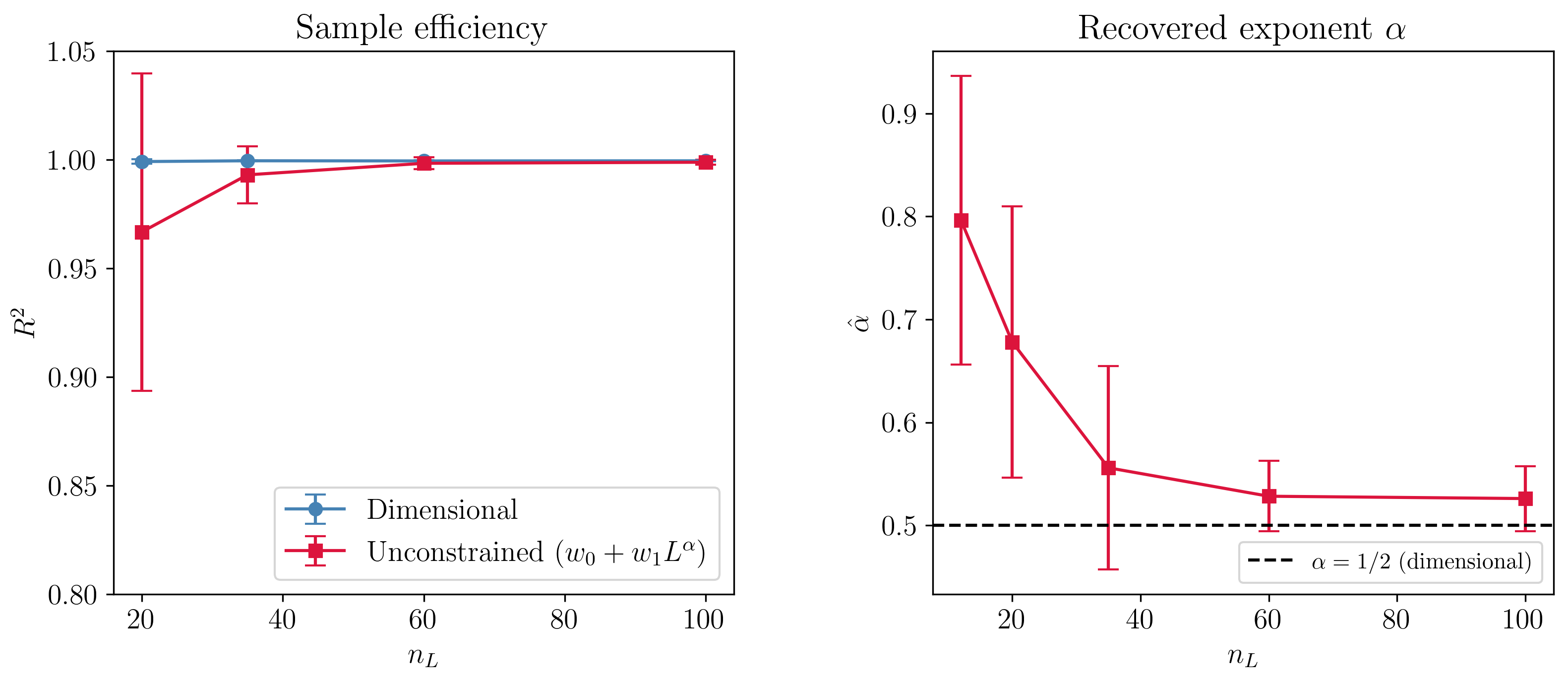}
  \caption{Dimensional vs.\ unconstrained power-law baseline on the simple pendulum. Left: out-of-sample $R^2$ as a function of $n_L$. Right: recovered exponent $\hat\alpha$ vs.\ $n_L$, skipping $n_L<20$; the dashed line marks the dimensional value $\alpha=1/2$. Error bars denote $\pm 1$ standard deviation over 30 independent trials. Both models fitted with \texttt{Minuit}.}
  \label{fig:unconstrained_baseline}
\end{figure}
}

\subsection{Black-body radiation}
\label{sec:blackbody_results}

As a second benchmark, we consider Planck's black-body law at fixed frequency.
Here the dimensional reduction still yields a one-dimensional invariant, but the
remaining dependence is transcendental; the benchmark therefore tests how
prefactor selection interacts with approximation error and noise.

\subsubsection{Physical setting and dimensional structure}

In frequency form, Planck's law for the spectral radiance of a black body is
\begin{align}
B_\nu(T)
=
\frac{2h\nu^3}{c^2}\,
\frac{1}{e^{h\nu/(k_B T)}-1},
\end{align}
where $\nu$ is the frequency, $T$ is the absolute temperature, $h$ is Planck's constant, $c$ is the speed of light, and $k_B$ is Boltzmann's constant. In the experiments below, $\nu$ is treated as a fixed physical parameter, and the dependence on temperature is learned from data. The variables are ordered as $(\nu, T, h, c, k_B)$, and the dimension matrix with respect to $\mathcal{B}=\{\mathsf{M},\mathsf{L},\mathsf{T},\mathsf{K}\}$ is:
\begin{align}
A=
\begin{pmatrix}
0 & 0 & 1 & 0 & 1 \\
0 & 0 & 2 & 1 & 2 \\
-1 & 0 & -1 & -1 & -2 \\
0 & 1 & 0 & 0 & -1
\end{pmatrix}.
\end{align}
Solving the homogeneous system $A\bm{\gamma}=\mathbf{0}$ yields a one-dimensional invariant space. A natural generator gives $\pi_1=\frac{k_B T}{h\nu}$. Thus, after dimensional reduction, the problem becomes one-dimensional. The admissible prefactors are obtained from the affine system $A\bm{\beta}=\mathbf{b}$. Setting $\bm{\beta} = \bm{\beta}_0 + \alpha\,\bm{\gamma}^{(1)}$ for the particular solution $\bm{\beta}_0=(3,0,1,-2,0)^\top$ and null-space vector $\bm{\gamma}^{(1)}=(-1,1,-1,0,1)^\top$, the admissible prefactor family becomes
\begin{align}
P_\alpha
=
\nu^{3-\alpha}T^\alpha h^{1-\alpha}c^{-2}k_B^\alpha.
\end{align}
The physically correct choice corresponds to $\alpha=0$, namely $P_0=\nu^3 h c^{-2}$, so the exact law can be written as
\begin{equation}
B_\nu
=
P_0\,
\Phi\!\left(\pi_1\right),
\label{eq:bb_factorization}
\end{equation}
with $\Phi(\pi_1)=\frac{2}{e^{1/\pi_1}-1}$. This benchmark is therefore especially informative for the prefactor scan. Any deviation from $\alpha=0$ can then be interpreted as evidence of finite-sample effects, dictionary bias, or noise sensitivity.

\subsubsection{Model and experimental regimes}

For this benchmark, the predictor is written as
\begin{equation}
\hat{B}_\nu
=
P_\alpha
\left[
w_0+\sum_{n=1}^{N_f}
\bigl(
w_n^{c}\cos(nu_1)+w_n^{s}\sin(nu_1)
\bigr)
\right],
\label{eq:bb_fourier_model}
\end{equation}
where $u_1$ is the rescaling of the invariant $\pi_1$ to the interval $[0,2\pi]$.

To examine separately the roles of prefactor selection and dictionary expressivity, we consider four regimes obtained by combining two prefactor-search ranges with two truncation orders:
\begin{itemize}
    \item \textbf{Regime A:} broad prefactor grid, $\alpha\in[-5,5]$, with low truncation $N_f=5$;
    \item \textbf{Regime B:} broad prefactor grid, $\alpha\in[-5,5]$, with high truncation $N_f=60$;
    \item \textbf{Regime C:} narrow prefactor grid, $\alpha\in[-0.5,0.5]$, with low truncation $N_f=5$;
    \item \textbf{Regime D:} narrow prefactor grid, $\alpha\in[-0.5,0.5]$, with high truncation $N_f=60$.
\end{itemize}
These four configurations make it possible to distinguish errors due to insufficient harmonic expressivity from errors due to imperfect identification of the dimensional scaling.

Synthetic datasets are generated by evaluating Planck's law over the temperature range
$T\in(0,10000]\ \mathrm{K}$,
at fixed frequency. The number of samples is varied from $50$ to $5000$ in ten steps. In the noisy experiments, additive Gaussian noise $\mathcal{N}(0,\,(\sigma_{\mathrm{rel}}\,\mathrm{std}(B_\nu))^2)$ is introduced at three relative levels, $\sigma_{\mathrm{rel}}\in\{0.02,\,0.2,\,0.6\}$, where $\mathrm{std}(B_\nu)$ is the standard deviation of the noiseless signal over the sampled temperature range.

\subsubsection{Noiseless results}

The noiseless results for the four regimes are reported in Tables~\ref{tab:bb_A}--\ref{tab:bb_D}. The main conclusion is that the correct dimensional prefactor is recovered.

\paragraph{Regime A: broad grid, low truncation.}
With a small dictionary, the model can fit the data well, but the prefactor scan may be biased for very small sample sizes. As shown in Table~\ref{tab:bb_A}, the method initially favors $\alpha=1$, and only for sufficiently large datasets does it converge to the correct value $\alpha=0$. This indicates that a rigid harmonic approximation can push part of the residual curvature into the dimensional prefactor.

\begin{table}[htbp]
    \caption{Black-body Regime A: broad prefactor grid ($\alpha\in[-5,5]$), low truncation ($N_f=5$).}
    \label{tab:bb_A}
    \centering
    \begin{tabular}{c c c c}
        \toprule
        $N_{\mathrm{dat}}$ & Best $\hat{\alpha}$ & Best $R^2$ & Best MSE \\
        \midrule
        50   & 1.0 & 0.999938 & 9.707071e-20 \\
        600  & 1.0 & 0.999970 & 4.605817e-20 \\
        1150 & 0.0 & 0.999998 & 3.589867e-21 \\
        1700 & 0.0 & 0.999998 & 3.440877e-21 \\
        2250 & 0.0 & 0.999998 & 3.339133e-21 \\
        2800 & 0.0 & 0.999998 & 3.255943e-21 \\
        3350 & 0.0 & 0.999998 & 3.182589e-21 \\
        3900 & 0.0 & 0.999998 & 3.115358e-21 \\
        4450 & 0.0 & 0.999998 & 3.051995e-21 \\
        5000 & 0.0 & 0.999998 & 2.992223e-21 \\
        \bottomrule
    \end{tabular}
\end{table}

\paragraph{Regime B: broad grid, high truncation.}
Increasing the truncation order largely removes that bias. As shown in Table~\ref{tab:bb_B}, the correct value $\alpha=0$ is recovered already at moderate sample size, and the resulting fit is essentially exact throughout the range. This confirms that sufficient expressivity in the dimensionless modulation helps disentangle functional approximation error from dimensional scaling.

\begin{table}[htbp]
    \caption{Black-body Regime B: broad prefactor grid ($\alpha\in[-5,5]$), high truncation ($N_f=60$).}
    \label{tab:bb_B}
    \centering
    \begin{tabular}{c c c c}
        \toprule
        $N_{\mathrm{dat}}$ & Best $\hat{\alpha}$ & Best $R^2$ & Best MSE \\
        \midrule
        50   & 0.5 & 1.0 & 6.046916e-28 \\
        600  & 0.0 & 1.0 & 2.807961e-28 \\
        1150 & 0.0 & 1.0 & 2.365705e-28 \\
        1700 & 0.0 & 1.0 & 2.157930e-28 \\
        2250 & 0.0 & 1.0 & 2.016073e-28 \\
        2800 & 0.0 & 1.0 & 1.906302e-28 \\
        3350 & 0.0 & 1.0 & 1.815028e-28 \\
        3900 & 0.0 & 1.0 & 1.737311e-28 \\
        4450 & 0.0 & 1.0 & 1.669088e-28 \\
        5000 & 0.0 & 1.0 & 1.608124e-28 \\
        \bottomrule
    \end{tabular}
\end{table}

\paragraph{Regime C: narrow grid, low truncation.}
When both the prefactor range and the harmonic basis are restricted, small but persistent deviations from the theoretical scaling remain visible; see Table~\ref{tab:bb_C}. In particular, the scan tends to settle near $\alpha=0.05$ for larger sample sizes. This suggests that when the dictionary is too rigid, the optimization may partially compensate by distorting the prefactor.

\begin{table}[htbp]
    \caption{Black-body Regime C: narrow prefactor grid ($\alpha\in[-0.5,0.5]$), low truncation ($N_f=5$).}
    \label{tab:bb_C}
    \centering
    \begin{tabular}{c c c c}
        \toprule
        $N_{\mathrm{dat}}$ & Best $\hat{\alpha}$ & Best $R^2$ & Best MSE \\
        \midrule
        50   & 0.05  & 0.999879 & 1.894525e-19 \\
        600  & -0.05 & 0.999977 & 3.485987e-20 \\
        1150 & 0.00  & 0.999998 & 3.589867e-21 \\
        1700 & 0.00  & 0.999998 & 3.440877e-21 \\
        2250 & 0.00  & 0.999998 & 3.339133e-21 \\
        2800 & 0.00  & 0.999998 & 3.255943e-21 \\
        3350 & 0.00  & 0.999998 & 3.182589e-21 \\
        3900 & 0.05  & 0.999998 & 2.954565e-21 \\
        4450 & 0.05  & 0.999998 & 2.900151e-21 \\
        5000 & 0.05  & 0.999998 & 2.847741e-21 \\
        \bottomrule
    \end{tabular}
\end{table}

\paragraph{Regime D: narrow grid, high truncation.}
With a richer harmonic basis, the prefactor again stabilizes close to the theoretical value; see Table~\ref{tab:bb_D}. Small oscillations around $\alpha=0$ remain for some intermediate sample sizes, but they correspond to numerically indistinguishable fits and do not affect the overall conclusion.

\begin{table}[htbp]
    \caption{Black-body Regime D: narrow prefactor grid ($\alpha\in[-0.5,0.5]$), high truncation ($N_f=60$).}
    \label{tab:bb_D}
    \centering
    \begin{tabular}{c c c c}
        \toprule
        $N_{\mathrm{dat}}$ & Best $\hat{\alpha}$ & Best $R^2$ & Best MSE \\
        \midrule
        50   & 0.50  & 1.0 & 6.046916e-28 \\
        600  & 0.00  & 1.0 & 2.807961e-28 \\
        1150 & 0.00  & 1.0 & 2.365705e-28 \\
        1700 & -0.10 & 1.0 & 2.033666e-28 \\
        2250 & -0.05 & 1.0 & 1.958330e-28 \\
        2800 & 0.00  & 1.0 & 1.906302e-28 \\
        3350 & 0.00  & 1.0 & 1.815028e-28 \\
        3900 & 0.00  & 1.0 & 1.737311e-28 \\
        4450 & 0.00  & 1.0 & 1.669088e-28 \\
        5000 & 0.00  & 1.0 & 1.608124e-28 \\
        \bottomrule
    \end{tabular}
\end{table}

Taken together, the noiseless experiments show that prefactor recovery is highly robust provided that the harmonic dictionary is expressive enough. When the dictionary is too restrictive, part of the residual functional error leaks into the prefactor scan.

\subsubsection{Results with noise}

We now repeat the same four regimes under additive Gaussian noise with levels $\sigma_{\mathrm{rel}}\in\{0.02,\,0.2,\,0.6\}$. The full numerical results are summarized in Figure~\ref{fig:bb_noise_summary} of Appendix~\ref{app:noisePlanck}. The main conclusion is that the dimensional component of the model remains substantially more stable than the functional one.

Across all regimes, the recovered exponent stays at or very near the theoretical value $\alpha=0$, even when the quality of the fitted modulation degrades at high noise. The broad-grid regimes recover $\alpha=0$ almost uniformly, while the narrow-grid regimes display only small deviations of order $\pm 0.05$ or $\pm 0.10$. This indicates that the dimensional skeleton acts as a robust low-complexity structural constraint, much less sensitive to noise than the detailed harmonic content of the fitted function.

A second clear trend is the expected bias-variance trade-off associated with the truncation order. Low-order dictionaries are more robust but may introduce approximation bias, which can then distort the estimated prefactor. High-order dictionaries remove most of that bias but may begin to overfit noise. This is most visible in Regime~B at $\sigma_{\mathrm{rel}}=0.6$, where the correct prefactor is still recovered but the fitted modulation starts to track noise fluctuations.

\begin{figure}[htbp]
  \centering
  \includegraphics[width=1\linewidth]{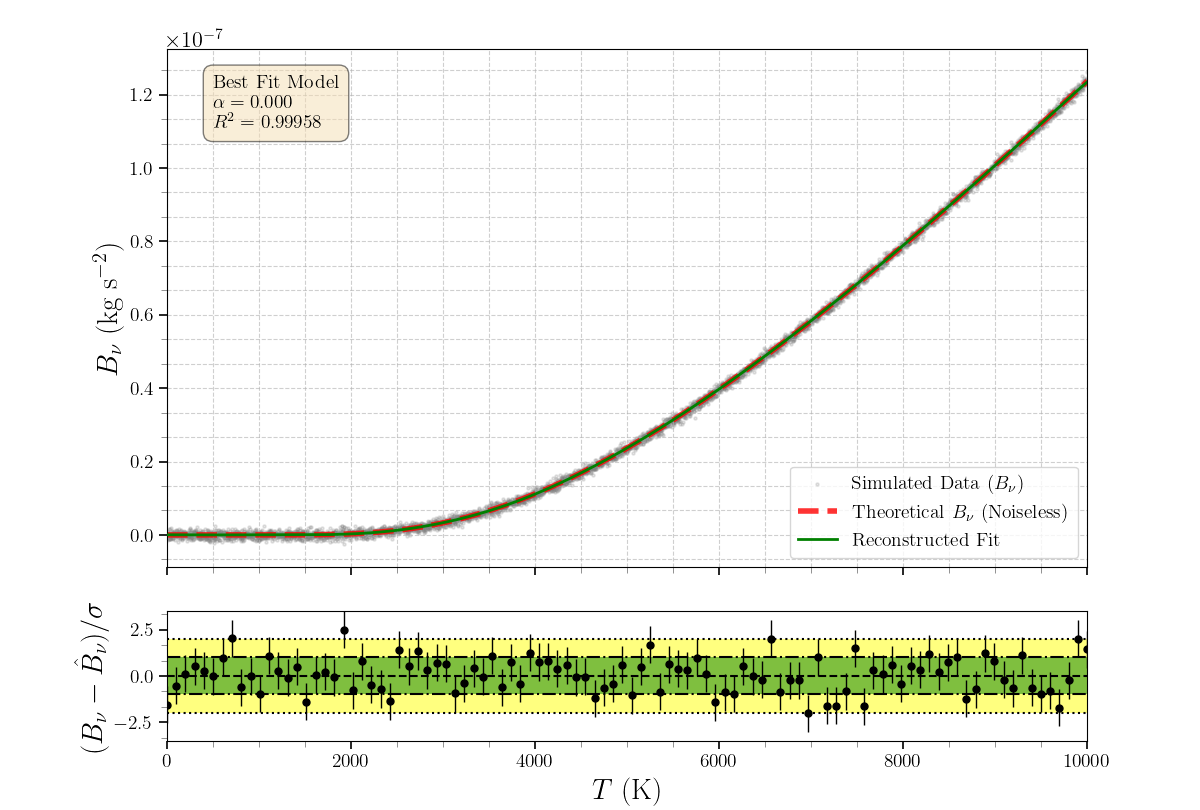}
  \caption{Regime~B ($N_f=60$, $\alpha\in[-5,5]$), low noise ($\sigma_{\mathrm{rel}}=0.02$, $N_{\text{dat}}=5000$): near-perfect reconstruction of the Planck curve.}
  \label{fig:bb_lownoise}
\end{figure}

\begin{figure}[htbp]
\centering
\includegraphics[width=1\textwidth]{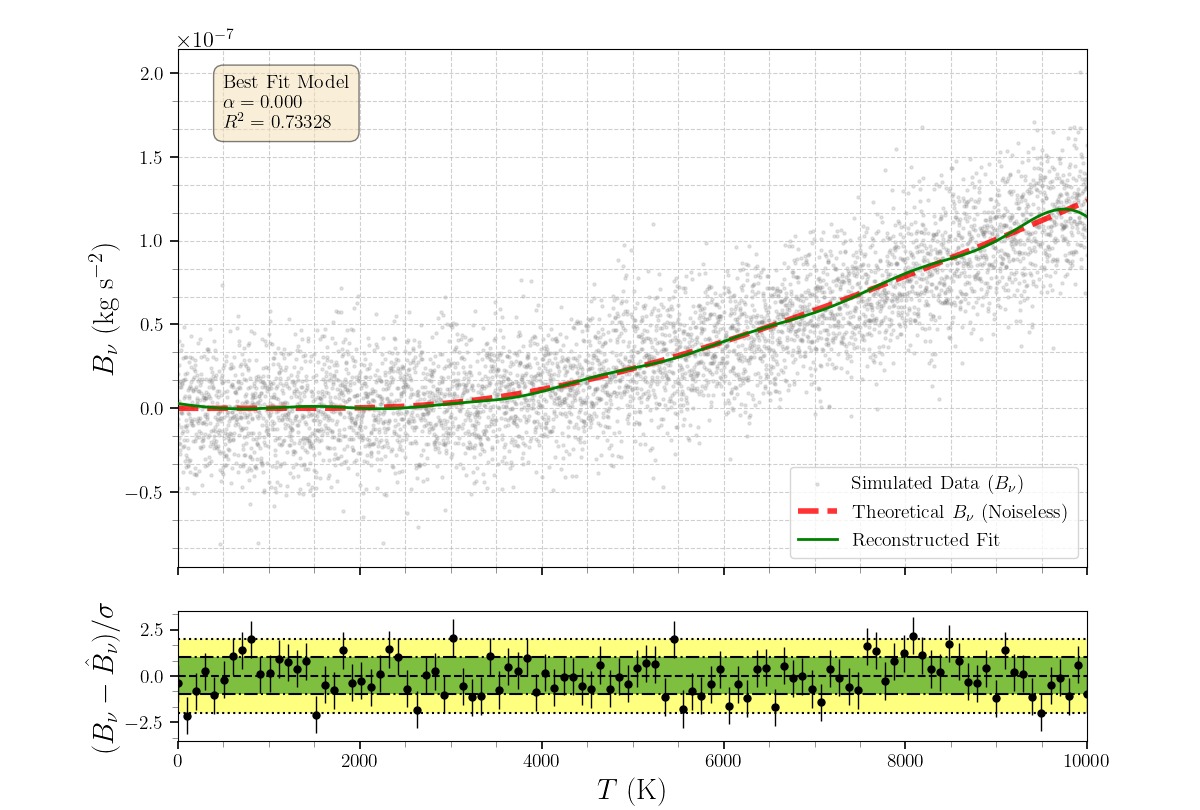}
\caption{Example of functional overfitting in the high-noise regime (Regime~B, $\sigma_{\mathrm{rel}}=0.6$, $N_f=60$). The prefactor scan still identifies the correct dimensional scaling, but the fitted modulation begins to track noise fluctuations.}
\label{fig:bb_overfit}
\end{figure}

Another useful diagnostic is the saturation of out-of-sample $R^2$ at moderate and high noise levels. Writing the noisy target as $y=y_{\mathrm{true}}+\varepsilon$ with $\varepsilon\sim\mathcal{N}(0,\sigma_{\mathrm{rel}}^2\,\mathrm{var}(y_{\mathrm{true}}))$, and normalizing the noiseless signal to unit variance so that $\mathrm{var}(y_{\mathrm{true}})=1$, even a perfect predictor of $y_{\mathrm{true}}$ is bounded by
\begin{equation}
R^2_\infty=\frac{1}{1+\sigma_{\mathrm{rel}}^2}.
\label{eq:r2_ceiling}
\end{equation}
{This saturation value assumes that the fitted model is an unbiased predictor of the noiseless signal. Therefore, systematic deviations below $R^2_\infty$ should not be attributed solely to observation noise; they also indicate approximation bias, model misspecification, or an insufficiently expressive dictionary.}
This gives $R^2_\infty\approx 0.999$ for $\sigma_{\mathrm{rel}}=0.02$, and $R^2_\infty\approx 0.735$ for $\sigma_{\mathrm{rel}}=0.6$, which agrees closely with the empirical plateaus seen in the noisy experiments. See figures~\ref{fig:bb_lownoise} and \ref{fig:bb_overfit}.

\subsubsection{Interpretation}

The black-body benchmark illustrates a useful distinction between the dimensional and functional parts of the model. The dimensional skeleton is recovered more robustly than the detailed modulation, even when the latter is only approximately captured or partially overfitted.

At the same time, the experiments show that prefactor selection and functional approximation are not fully separable. When the harmonic basis is too rigid, approximation bias can leak into the prefactor scan, producing small but systematic deviations from the theoretical scaling. When the basis is sufficiently expressive, the scan stabilizes near the correct value and the remaining error is concentrated in the dimensionless fit itself.

For this reason, the black-body benchmark provides a more stringent validation of the framework than the simple pendulum. It shows that dimensional consistency remains informative and robust even when the residual law is highly non-polynomial, while also making clear that the success of prefactor recovery depends on matching the dictionary complexity to the structure of the dimensionless modulation.

{
\subsubsection{Validation on COBE/FIRAS experimental data}

We validate the framework on real experimental data: the cosmic microwave background (CMB) monopole spectrum measured by the FIRAS instrument on board the COBE satellite~\cite{fixsen1996}. The dataset provides the spectral radiance $B_\nu$ at 43 frequency points in the range $\nu\in[68, 639]$\,GHz, at the fixed CMB temperature $T=2.725$\,K. The dimensional structure is constructed exactly as in the synthetic benchmark: the invariant is $\pi_1=k_B T/h\nu$ and the admissible prefactor family is $P_\alpha=\nu^{3-\alpha}\,T^\alpha\,h^{1-\alpha}\,c^{-2}\,k_B^\alpha$. We use \texttt{RidgeCV} with 5-fold cross-validation. The single-fit validation shown in Figure~\ref{fig:firas_fit} uses $N_f=30$ and a 30/13 train--test split, approximately a 70/30 split. For the bootstrap analysis in Figure~\ref{fig:firas_bagging}, we use $N_f=8$ harmonic terms. This lower truncation remains sufficiently expressive for the FIRAS spectrum while reducing overfitting artifacts under resampling, which is important given the limited number of data points.

A single train/test split selects $\hat\alpha=0.4$ with $R^2_{\text{test}}=0.999998$; the theoretical $\alpha=0$ yields $R^2_{\text{test}}=0.999982$ on the same split, the difference is negligible ($\Delta R^2\approx 10^{-5}$). To assess whether $\hat\alpha=0.4$ is a statistical fluctuation, we repeat the experiment over 500 classic bootstrap samples (with replacement), training each on the in-bag observations and testing on the out-of-bag points (Figure~\ref{fig:firas_bagging}). The bootstrap distribution yields $\hat\alpha = 0.15 \pm 0.36$, concentrated near zero over the scan range $[-1,1]$, confirming that the prefactor exponent $\alpha$ is compatible with zero and is not sharply resolved from only 43 data points at fixed temperature. Furthermore, $\alpha=0$ produces a mean out-of-bag $R^2$ of $0.976$ across bootstrap samples, higher than the $R^2=0.947$ achieved by the sample-selected $\hat\alpha$. This suggests that the data are too scarce to resolve the prefactor exponent sharply, and that additional measurements would likely concentrate the estimate more strongly around $\hat\alpha=0$. The fit for $\alpha=0$ is shown in Figure~\ref{fig:firas_fit}.

\begin{figure}[htbp]
  \centering
  \includegraphics[width=1\textwidth]{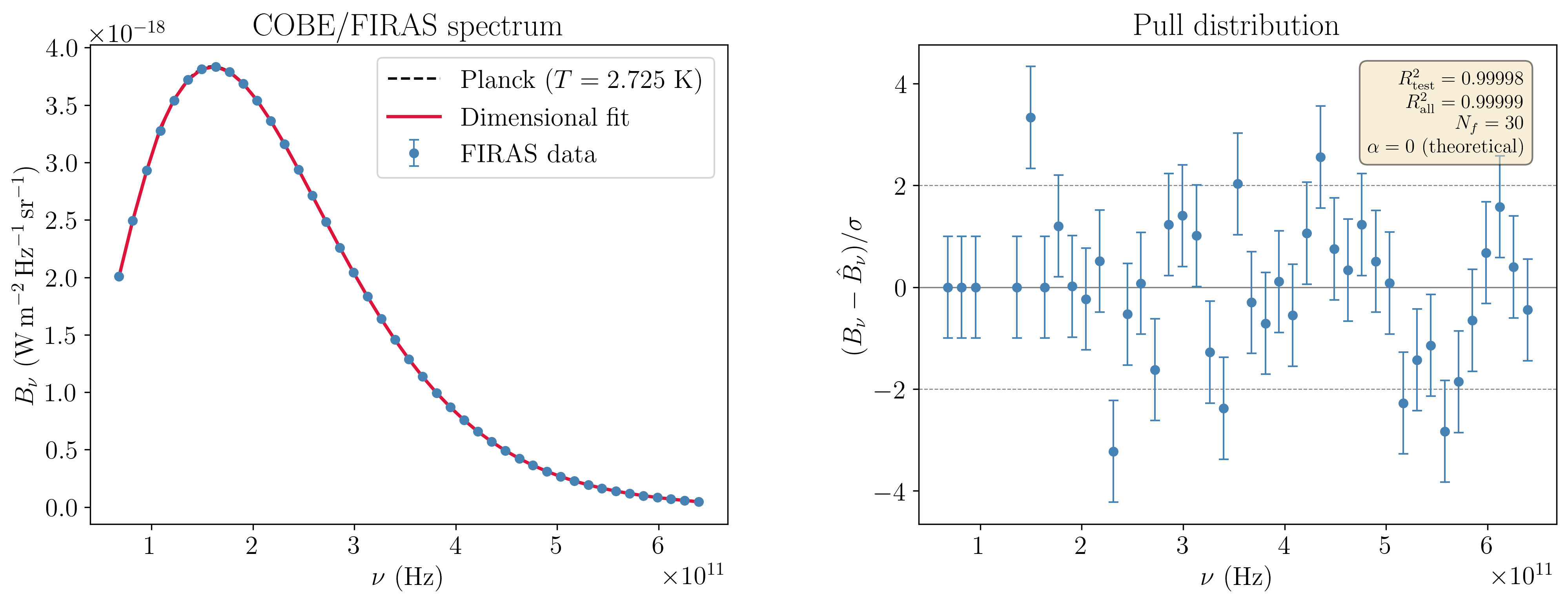}
  \caption{Fit of the dimensionally constrained model to COBE/FIRAS CMB monopole spectrum data (43 points, $\alpha=0$, $N_f=30$). Left: spectral radiance $B_\nu$ as a function of frequency; the theoretical Planck curve at $T=2.725$\,K (dashed) is overlaid with the FIRAS measurements (blue circles) and the dimensional fit (red line). Right: pull distribution $(B_\nu-\hat{B}_\nu)/\sigma$ in units of the FIRAS measurement uncertainty $\sigma$.}
  \label{fig:firas_fit}
\end{figure}

\begin{figure}[htbp]
  \centering
  \includegraphics[width=1\textwidth]{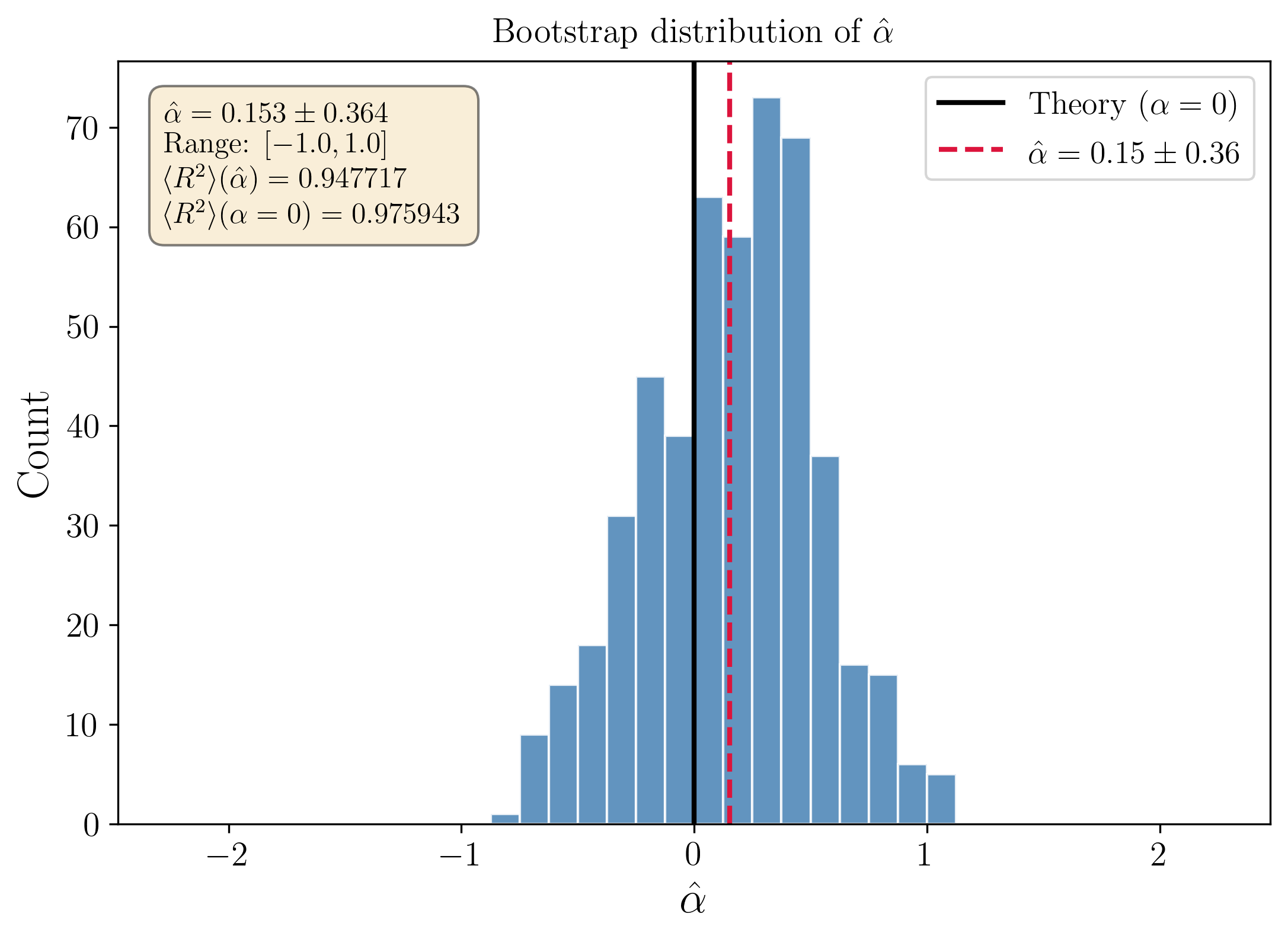}
  \caption{Bootstrap distribution of the recovered prefactor exponent $\hat\alpha$ over 500 classic bootstrap samples (with replacement) of the 43 FIRAS data points, using $N_f=8$ harmonic terms. The theoretical value $\alpha=0$ (solid black line) and the bootstrap result $\hat\alpha=0.15\pm0.36$ (dashed red line) are indicated. The distribution is concentrated near zero with a modest spread ($\sigma\approx0.36$).}
  \label{fig:firas_bagging}
\end{figure}
}

{
\subsubsection{Harmonic versus Chebyshev basis}

The choice of harmonic expansions for the dimensionless modulation deserves justification, since other bases such as Chebyshev polynomials, splines, or radial basis functions could also be employed. We compare the harmonic basis against Chebyshev polynomials mapped to $[-1,1]$ on the black-body benchmark, using the same number of basis functions and the same \texttt{RidgeCV} pipeline. Figure~\ref{fig:cheb_bb} shows the results.

With $N_f=3$ to $N_f=50$ harmonic terms, the out-of-sample $R^2$ remains stable at $0.9996$, while the Chebyshev basis degrades from $0.9997$ at low order to $0.866$ at $N_f=50$. The Chebyshev expansion suffers from Runge's phenomenon when approximating the transcendental modulation $\Phi(\pi_1)=2/(e^{1/\pi_1}-1)$: high-order algebraic polynomials oscillate strongly near the boundaries of the domain, whereas the harmonic expansion captures the exponential structure with a smooth spectral basis. Furthermore, Chebyshev requires substantially more training samples to generalize: with 20 to 30 training points, the harmonic model attains $R^2>0.999$ while Chebyshev remains below $0.82$.

\begin{figure}[htbp]
  \centering
  \includegraphics[width=0.98\textwidth]{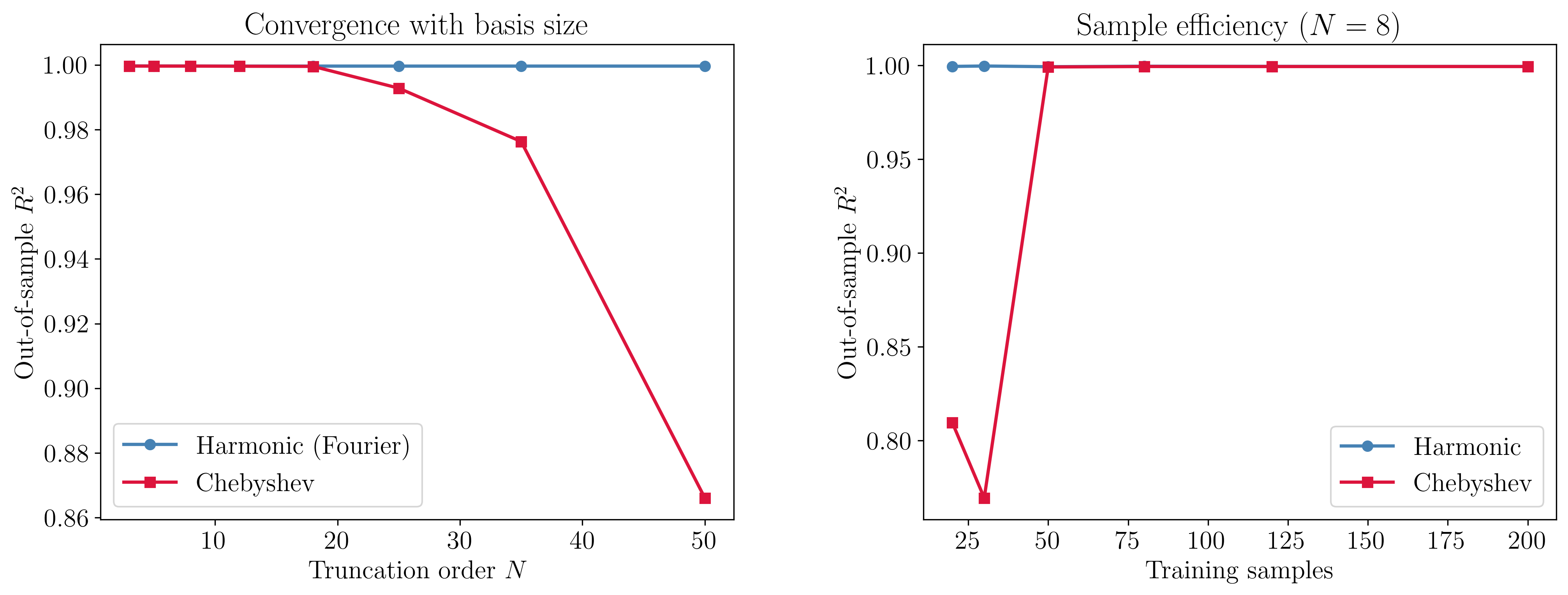}
  \caption{Comparison of harmonic and Chebyshev polynomial bases on the black-body benchmark. Left: out-of-sample $R^2$ as a function of the truncation order $N_f$, with both bases using the same number of features ($1+2N_f$). Right: sample efficiency at fixed $N_f=8$, showing that the harmonic expansion generalizes reliably from as few as 20 training points while Chebyshev requires at least 100.}
  \label{fig:cheb_bb}
\end{figure}
}

{
\subsubsection{Comparison with Villar et al.\ (2023)}

We compare our harmonic-based approach with a Villar et al.--style baseline inspired by the units-equivariant framework of Villar et al.~\cite{villar2023units}. In this baseline, dimensional analysis is used to construct dimensionless inputs, and inference is then performed in the resulting dimensionless space using a standard machine-learning model. Both approaches share the same dimensional reduction: the target is divided by the dimensional prefactor $P_0=\nu^3 h/c^2$, and the regression is performed on the invariant $\pi_1=k_B T/h\nu$. Our method uses a harmonic expansion with $N_f=8$ terms and \texttt{RidgeCV} with cross-validation, yielding a linear-in-parameters predictor with explicit harmonic coefficients. The Villar-style baseline employs a multi-layer perceptron (MLP) with three hidden layers of 128, 128, and 64 units and ReLU activations, trained on the same invariant coordinate.

The results are summarized in Table~\ref{tab:villar_comparison} and Figure~\ref{fig:villar_comparison}. Both methods achieve high accuracy, but the harmonic expansion consistently outperforms the MLP, particularly at low sample sizes ($R^2=0.999996$ vs.\ $0.989688$ at $N=30$ training points) and under noise (e.g., $R^2=0.721$ vs.\ $0.683$ at $\sigma_{\text{rel}}=0.6$). The harmonic model uses a convex fitting objective and explicit coefficients for the dimensionless response. The MLP is more flexible, but in this experiment it requires more data to generalize and involves a non-convex optimization problem.

\begin{table}[htbp]
\centering
\caption{Comparison between the proposed harmonic model and a Villar et al.--style MLP baseline on the black-body benchmark ($N_f=8$, 70/30 train/test split).}
\label{tab:villar_comparison}
\footnotesize
\setlength{\tabcolsep}{8pt}
\renewcommand{\arraystretch}{1.08}
\begin{tabular}{l c c}
\toprule
$\sigma_{\text{rel}}$ & Harmonic (ours) & MLP on $\pi_1$ (Villar-style baseline) \\
\midrule
0.00 & 1.000000 & 0.987803 \\
0.02 & 0.999652 & 0.987709 \\
0.10 & 0.990905 & 0.980457 \\
0.30 & 0.923498 & 0.891541 \\
0.60 & 0.721088 & 0.683265 \\
\bottomrule
\end{tabular}
\end{table}

\begin{figure}[htbp]
  \centering
  \includegraphics[width=0.98\textwidth]{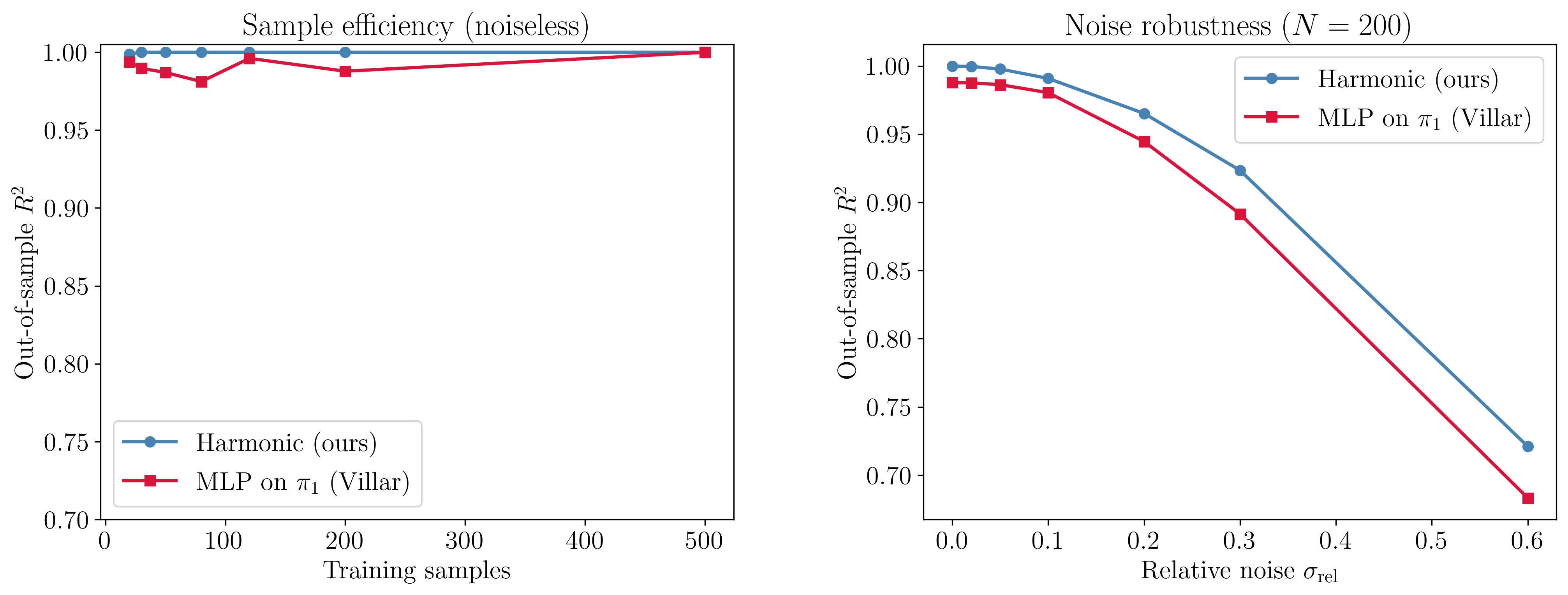}
  \caption{Comparison of our harmonic expansion method against a Villar et al.\-style MLP operating on the same dimensionless invariant space. Left: out-of-sample $R^2$ as a function of training dataset size (noiseless). Right: out-of-sample $R^2$ under increasing additive noise ($N=200$). The harmonic model matches or exceeds the MLP in these tests while using a linear fit in the coefficients.}
  \label{fig:villar_comparison}
\end{figure}
}

This comparison should not be interpreted as a replacement for the more general
units-equivariant framework of Villar et al., but as a controlled baseline
within the same dimensionless input space. The purpose is to isolate the effect
of replacing a flexible black-box regressor by an explicit harmonic dictionary.
In this setting, the harmonic model keeps the same dimensional reduction but uses a smaller explicit dictionary instead of a neural regressor.

{
Taken together, these comparisons provide complementary ablations of the
proposed framework. The unconstrained pendulum baseline removes the dimensional
reduction and fits the raw variables directly, testing the effect of imposing
Buckingham structure and an admissible dimensional prefactor. The Chebyshev
comparison keeps the same dimensional prefactor and invariant coordinate but
replaces the harmonic dictionary by a non-periodic polynomial basis. Finally, the Villar et al.--style baseline keeps
the same dimensionless input space but replaces the explicit harmonic dictionary
by a flexible black-box regressor. Together, these comparisons separate the effects of dimensional reduction, prefactor structure, dictionary choice, and regression architecture.
}

\subsection{Double pendulum Lyapunov field}
\label{sec:doublependulum_results}

As a third and substantially more demanding benchmark, we consider the maximal
Lyapunov exponent field of the double pendulum. The target is a structured
two-dimensional response surface with strong anisotropy and sharp transitions,
which makes it a stringent test of dictionary design on invariant space.

\subsubsection{Physical setting and dimensional structure}

The double pendulum consists of two point masses $m_1$ and $m_2$ connected by rigid massless rods of lengths $l_1$ and $l_2$ in a uniform gravitational field $g$. Its angular equations of motion are nonlinear and strongly coupled:
\begin{small}
\begin{align}
  \ddot{\theta}_1 &=
  \frac{
    -g(2m_1+m_2)\sin\theta_1
    - m_2 g\sin(\theta_1-2\theta_2)
    - 2m_2\sin(\theta_1-\theta_2)
    \bigl(\dot\theta_2^2 l_2+\dot\theta_1^2 l_1\cos(\theta_1-\theta_2)\bigr)
  }{
    l_1\bigl(2m_1+m_2-m_2\cos(2\theta_1-2\theta_2)\bigr)
  },\nonumber\\
  \ddot{\theta}_2 &=
  \frac{
    2\sin(\theta_1-\theta_2)\bigl(
      \dot\theta_1^2 l_1(m_1+m_2)
      + g(m_1+m_2)\cos\theta_1
      + \dot\theta_2^2 l_2 m_2\cos(\theta_1-\theta_2)
    \bigr)
  }{
    l_2\bigl(2m_1+m_2-m_2\cos(2\theta_1-2\theta_2)\bigr)
  }.
\end{align}
\end{small}
To quantify sensitivity to initial conditions, we study the maximal Lyapunov exponent
\begin{align}
\lambda_{\text{max}}
=
\lim_{t\to\infty}\lim_{\|\delta_0\|\to 0}
\frac{1}{t}\log\frac{\|\delta(t)\|}{\|\delta_0\|}.
\end{align}
Numerically, $\lambda_{\text{max}}$ is estimated from the divergence of nearby trajectories initialized with a small angular perturbation. The resulting map
$(\theta_1,\theta_2)\mapsto \lambda_{\text{max}}(\theta_1,\theta_2)$ defines a structured response surface over the angular configuration space. We use the dimensional basis $\mathcal{B}=\{\mathsf{M},\mathsf{L},\mathsf{T}\}$. The dimension matrix for the dimensional variables $(m_1,m_2,l_1,l_2,g)$ is
\begin{align}
A=
\begin{pmatrix}
1 & 1 & 0 & 0 & 0 \\
0 & 0 & 1 & 1 & 1 \\
0 & 0 & 0 & 0 & -2
\end{pmatrix}.
\end{align}
Since the angular variables are already dimensionless, Buckingham's theorem gives $N_\Pi = 7-3 = 4$ independent invariants when the full variable set $(m_1,m_2,l_1,l_2,g,\theta_1,\theta_2)$ is considered. Solving $A\bm{\gamma}=\mathbf{0}$ yields, 
\begin{align}
\pi_1=\theta_1,
\qquad
\pi_2=\theta_2,
\qquad
\pi_3=\frac{m_2}{m_1},
\qquad
\pi_4=\frac{l_2}{l_1}.
\end{align}
For the target signature $[\lambda_{\text{max}}]=T^{-1}$, solving $A\bm{\beta}=\mathbf{b}$ yields the admissible prefactor family
\begin{align}
P_{\eta,\xi}
=
\left(\frac{m_2}{m_1}\right)^\eta
\left(\frac{l_2}{l_1}\right)^{-\xi}
\sqrt{\frac{g}{l_1}}.
\end{align}
In the symmetric baseline configuration, where $m_1=m_2$ and $l_1=l_2$, this reduces to the unique minimal prefactor $P=\sqrt{\frac{g}{l_1}}$. Accordingly, the remaining learning task is concentrated entirely in the two-dimensional angular modulation.

\subsubsection{Data generation and benchmark configurations}

Two numerical configurations are considered.

\begin{itemize}
    \item In the \emph{symmetric baseline case}, we fix $m_1=m_2=1\,\mathrm{kg}$, $l_1=l_2=200\,\mathrm{m}$, and sample the Lyapunov field on a $401\times 401$ grid over $(\theta_1,\theta_2)\in[-\pi,\pi]^2$.
\item In the \emph{asymmetric case}, unequal masses are used and the field is sampled on a $201\times 201$ grid, while the integration protocol remains unchanged. This second configuration is included to verify that the advantage of the separable dictionary is not specific to the symmetric geometry of the baseline field.
\end{itemize}
For both settings, the learned model has the generic form
\begin{equation}
\hat{\lambda}_{\text{max}}
=
P_{\eta,\xi}\,\Phi(\theta_1,\theta_2),
\label{eq:doublependulum_factorization}
\end{equation}
with $P=\sqrt{g/l_1}$ in the symmetric case. Since the nontrivial dependence is genuinely two-dimensional, this benchmark is the natural place to compare the two harmonic constructions introduced earlier.

\subsubsection{Phase-combination dictionary}

We first use the phase-combination basis
\begin{equation}
\Phi(u_1,u_2) = w_0 + \sum_{\mathbf{n}\neq\mathbf{0}} \Bigl[ w_{\mathbf{n}}^{c}\cos(n_1u_1+n_2u_2) + w_{\mathbf{n}}^{s}\sin(n_1u_1+n_2u_2) \Bigr],
\label{eq:doublependulum_phase_dictionary}
\end{equation}
where $(u_1,u_2)$ are the rescaled angular invariants. This basis is straightforward to implement, but it couples both directions through a single phase and therefore has limited flexibility when the response surface exhibits different structures along different angular directions. The corresponding performance in the symmetric baseline configuration is summarized in Table~\ref{tab:lyapunov_phase}. Throughout this benchmark, $N_f$ denotes the per-coordinate truncation order used uniformly across both angular invariant directions (i.e.\ $N_f^{(1)}=N_f^{(2)}=N_f$).

\begin{table}[htbp]
    \caption{Performance of the phase-combination dictionary for the symmetric baseline configuration ($m_1=m_2$, $l_1=l_2$).}
    \label{tab:lyapunov_phase}
    \centering
    \begin{tabular}{c c c}
        \toprule
        Truncation ($N_f$) & $R^2$ & MSE \\
        \midrule
        5  & 0.791778 & 0.517102 \\
        20 & 0.816322 & 0.456150 \\
        30 & 0.823415 & 0.438534 \\
        \bottomrule
    \end{tabular}
\end{table}

The improvement with truncation order is modest, and the fit saturates near $R^2\approx 0.82$. This suggests that the main limitation is not the number of frequencies alone, but the geometry of the basis itself.

\subsubsection{Separable tensor-product dictionary}

We next replace the phase-combination basis by the separable tensor-product dictionary
$\Phi(u_1,u_2)
=
\sum_{\mathbf{k}\in\mathcal{I}_{\mathbf{k}}} w_{\mathbf{k}}\,\Psi_{\mathbf{k}}(u_1,u_2)$, where each feature is a product of one-dimensional trigonometric factors in $u_1$ and $u_2$. This separates the two angular directions before interaction terms are introduced. The results for the symmetric baseline configuration are given in Table~\ref{tab:lyapunov_separable}.

\begin{table}[h!]
    \caption{Performance of the separable tensor-product dictionary for the symmetric baseline configuration ($m_1=m_2$, $l_1=l_2$).}
    \label{tab:lyapunov_separable}
    \centering
    \begin{tabular}{c c c}
        \toprule
        Truncation ($N_f$) & $R^2$ & MSE \\
        \midrule
        5  & 0.860839 & 0.345595 \\
        20 & 0.933313 & 0.165611 \\
        \bottomrule
    \end{tabular}
\end{table}

At the same nominal truncation level, the separable basis outperforms the phase-combination model. This indicates that the limitation of the simpler representation is geometric: the Lyapunov field is anisotropic, and the dictionary must resolve different directional structures. The ten largest fitted coefficients for the $N_f=20$ separable model are shown in Table~\ref{tab:lyapunov_coeffs}. For readability we write $C_{k_1,k_2}$ for the coefficient of $\cos(k_1 u_1)\cos(k_2 u_2)$ and $S_{k_1,k_2}$ for the coefficient of $\sin(k_1 u_1)\sin(k_2 u_2)$; each corresponds uniquely to a scalar weight $w_{\mathbf{k}}$ from the tensor-product expansion of Section~\ref{sec:learning_physical_laws}. The subscript pair $(k_1, k_2)$ gives the angular-frequency index along the $\theta_1$ and $\theta_2$ invariant directions, respectively. Only even-parity products survive in the parity-reduced model (Table~\ref{tab:lyapunov_coeffs_35}), so no mixed $\sin(k_1 u_1)\cos(k_2 u_2)$ or $\cos(k_1 u_1)\sin(k_2 u_2)$ terms appear.

\begin{table}[h!]
    \caption{Ten largest coefficients for the separable model at $N_f=20$.}
    \label{tab:lyapunov_coeffs}
    \centering
    \begin{tabular}{l c}
        \toprule
        Term & Value \\
        \midrule
        $ C_{1,0} $ (cos-cos) & -7.976002 \\
        $ S_{1,1} $ (sin-sin) & -3.137987 \\
        $ C_{0,1} $ (cos-cos) & -2.709306 \\
        $ S_{1,2} $ (sin-sin) & 2.003437 \\
        $ S_{3,2} $ (sin-sin) & 1.498303 \\
        $ C_{2,3} $ (cos-cos) & -1.259975 \\
        $ C_{2,1} $ (cos-cos) & 1.184035 \\
        $ C_{2,0} $ (cos-cos) & 1.168022 \\
        $ S_{1,3} $ (sin-sin) & -0.956398 \\
        $ C_{3,1} $ (cos-cos) & -0.908308 \\
        \bottomrule
    \end{tabular}
\end{table}

\subsubsection{Parity reduction}

Inspection of the fitted coefficients and of the Lyapunov field itself reveals a strong approximate symmetry under the joint reflection $(\theta_1,\theta_2)\mapsto(-\theta_1,-\theta_2)$. This motivates imposing the parity reduction, retaining only the even-parity blocks $\cos(k_1u_1)\cos(k_2u_2)$ and $\sin(k_1u_1)\sin(k_2u_2)$. The resulting performance is summarized in Table~\ref{tab:lyapunov_parity}.

\begin{table}[htbp]
    \caption{Performance of parity-reduced dictionaries for the symmetric baseline configuration ($m_1=m_2$, $l_1=l_2$).}
    \label{tab:lyapunov_parity}
    \centering
    \begin{tabular}{l c c c}
        \toprule
        Dictionary Type & $N_f$ & $R^2$ & MSE \\
        \midrule
        Separable + Parity & 20 & 0.933290 & 0.165668 \\
        Phase-Comb + Parity & 40 & 0.826657 & 0.430482 \\
        Phase-Comb + Parity & 50 & 0.828744 & 0.425301 \\
        Separable + Parity & 35 & 0.950272 & 0.123495 \\
        \bottomrule
    \end{tabular}
\end{table}

Parity reduction alone does not resolve the limitations of the phase-combination basis. The best result is obtained by the parity-reduced separable model, which reaches $R^2=0.950272$ at $N_f=35$. Thus, symmetry is useful only after the dictionary geometry is already well matched to the target field. For completeness, the ten largest coefficients of the best model are reported in Table~\ref{tab:lyapunov_coeffs_35}.

\begin{table}[htbp]
  \caption{Ten largest coefficients for the parity-reduced separable model at $N_f=35$.}
  \label{tab:lyapunov_coeffs_35}
  \centering
  \begin{tabular}{l c}
    \toprule
    Term & Value \\
    \midrule
    $C_{1,0}$ (cos-cos) & -7.972290 \\
    $S_{1,1}$ (sin-sin) & -3.137987 \\
    $C_{0,1}$ (cos-cos) & -2.708697 \\
    $S_{1,2}$ (sin-sin) &  2.003437 \\
    $S_{3,2}$ (sin-sin) &  1.498303 \\
    $C_{2,3}$ (cos-cos) & -1.260299 \\
    $C_{2,1}$ (cos-cos) &  1.187112 \\
    $C_{2,0}$ (cos-cos) &  1.163263 \\
    $S_{1,3}$ (sin-sin) & -0.956398 \\
    $C_{3,1}$ (cos-cos) & -0.910151 \\
    \bottomrule
  \end{tabular}
\end{table}

\begin{figure}[h!]
  \centering
  % First Row
  \begin{subfigure}[t]{0.48\textwidth}
    \centering
    \includegraphics[width=\textwidth]{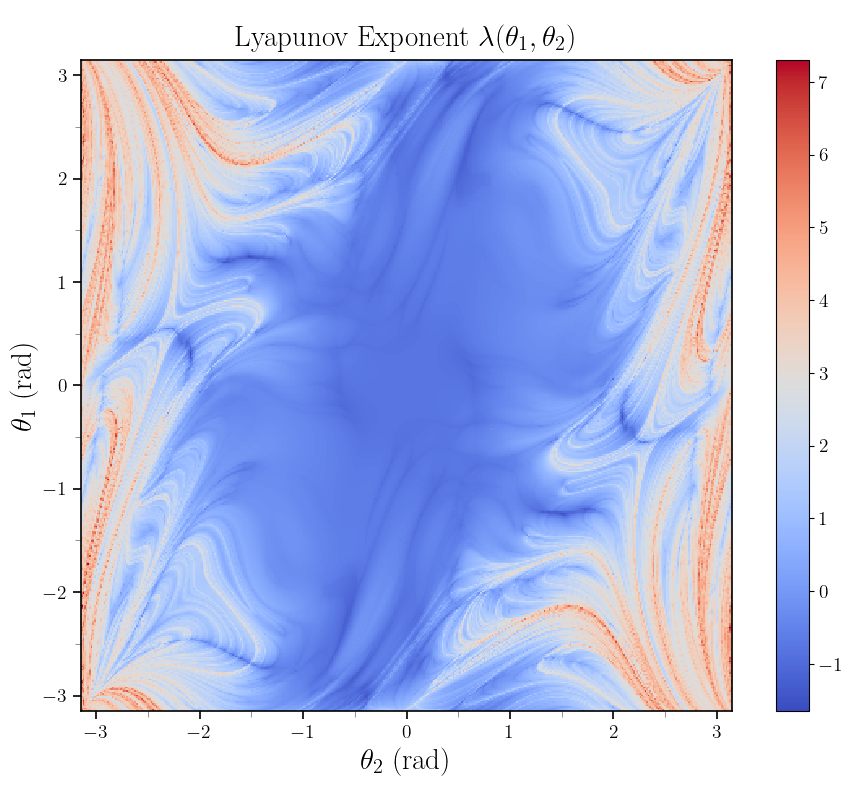}
    \caption{Ground-truth Lyapunov field.}
    \label{fig:lyap_truth}
  \end{subfigure}\hfill
  \begin{subfigure}[t]{0.48\textwidth}
    \centering
    \includegraphics[width=\textwidth]{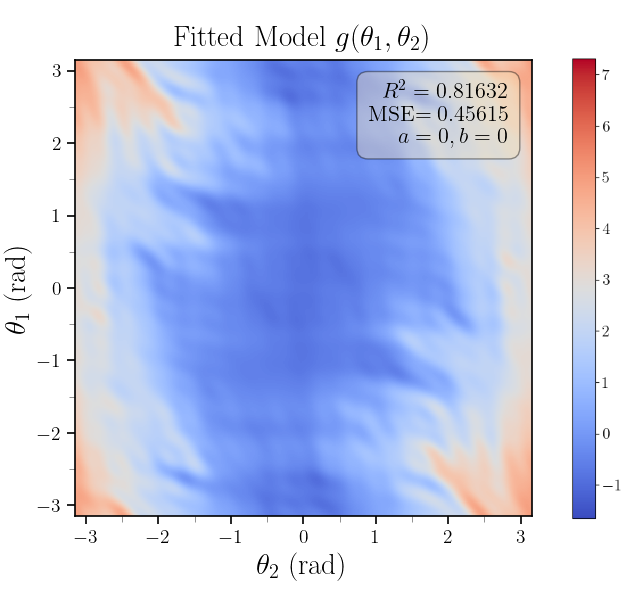}
    \caption{Phase-combination fit at $N_f=20$.}
    \label{fig:lyap_phase}
  \end{subfigure}

  \vspace{1em} % Adjust vertical spacing between rows

  % Second Row
  \begin{subfigure}[t]{0.48\textwidth}
    \centering
    \includegraphics[width=\textwidth]{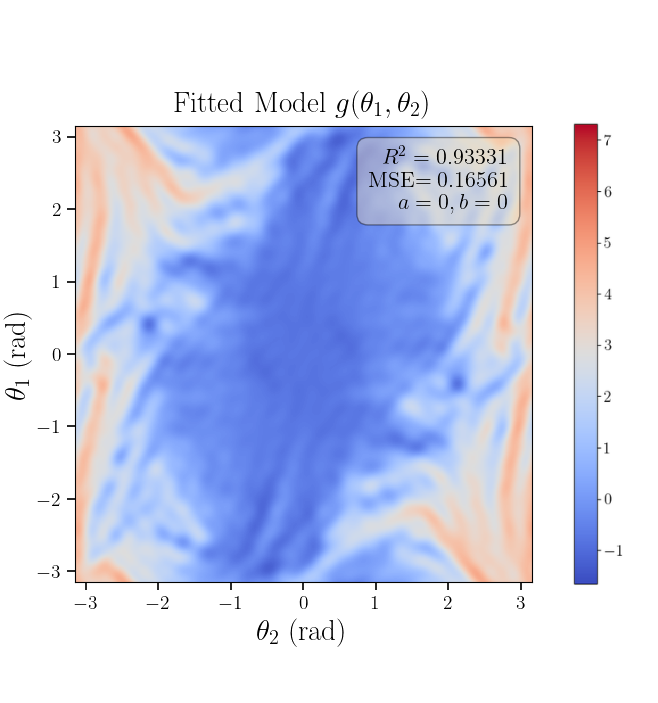}
    \caption{Separable fit at $N_f=20$.}
    \label{fig:lyap_sep}
  \end{subfigure}\hfill
  \begin{subfigure}[t]{0.48\textwidth}
    \centering
    \includegraphics[width=\textwidth]{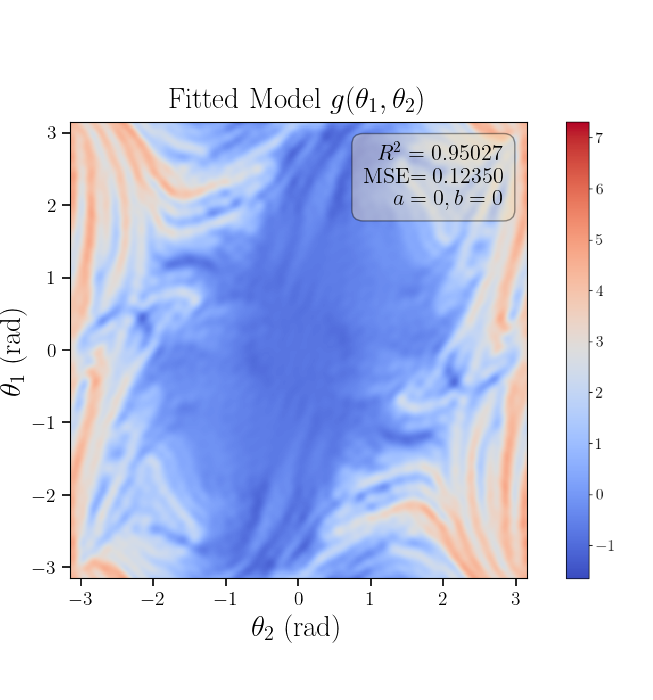}
    \caption{Parity-reduced separable fit at $N_f=35$.}
    \label{fig:lyap_parity}
  \end{subfigure}

  \caption{Progressive improvement in the reconstruction of the double-pendulum Lyapunov field. The separable tensor-product dictionary is essential for resolving the anisotropic angular structure, while parity reduction improves efficiency without degrading accuracy.}
  \label{fig:lyapunov_comparison}
\end{figure}

Figure~\ref{fig:lyapunov_comparison} shows the progression in reconstruction quality across the different dictionary choices.

\subsubsection{Asymmetric case}

We also consider an asymmetric mass configuration, for which the ratio invariant $m_2/m_1$ is no longer trivial. In this setting, the prefactor family
\begin{align}
P_{\eta,\xi}
=
\left(\frac{m_2}{m_1}\right)^\eta
\left(\frac{l_2}{l_1}\right)^{-\xi}
\sqrt{\frac{g}{l_1}}
\end{align}
contains nontrivial free parameters. Once the ratio invariants are fixed numerically, however, constant multiplicative factors are readily absorbed into the harmonic coefficients, so predictive performance depends much more strongly on the dictionary choice than on the precise prefactor exponent. The corresponding results are reported in Table~\ref{tab:lyapunov_asym}.

\begin{table}[htbp]
    \caption{Results for the asymmetric-mass double-pendulum configuration.}
    \label{tab:lyapunov_asym}
    \centering
    \begin{tabular}{l c c c}
        \toprule
        Dictionary Type & $N_f$ & $R^2$ & MSE \\
        \midrule
        Phase-Combination & 5  & 0.704248 & 3.395539 \\
        Phase-Combination & 20 & 0.717002 & 3.249113 \\
        Separable + Parity & 5  & 0.908818 & 1.046867 \\
        Separable + Parity & 20 & 0.942478 & 0.660411 \\
        \bottomrule
    \end{tabular}
\end{table}

\begin{figure}[h!]
  \centering
  % First Row
  \begin{subfigure}[t]{0.48\textwidth}
    \centering
    \includegraphics[width=\textwidth]{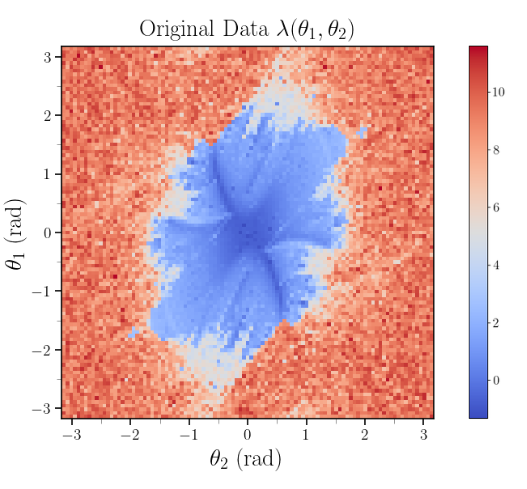}
    \caption{Ground-truth Lyapunov field.}
    \label{fig:lyap_asym_phase_5}
  \end{subfigure}\hfill
  \begin{subfigure}[t]{0.48\textwidth}
    \centering
    \includegraphics[width=\textwidth]{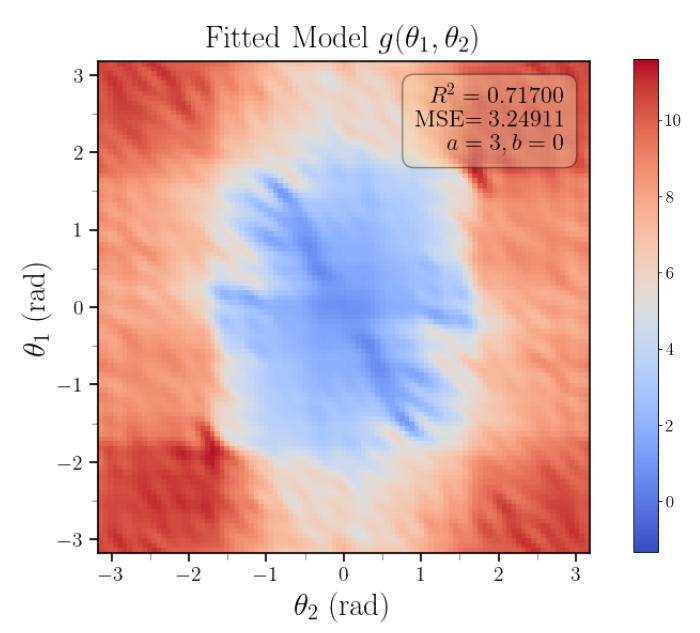}
    \caption{Phase-combination fit at $N_f=20$.}
    \label{fig:lyap_asym_phase_20}
  \end{subfigure}
  \vspace{1em} 
  % Second Row
  \begin{subfigure}[t]{0.48\textwidth}
    \centering
    \includegraphics[width=\textwidth]{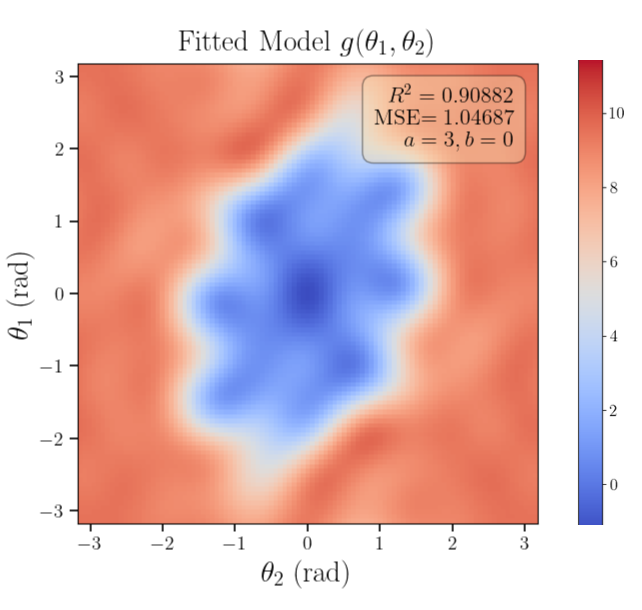}
    \caption{Parity-reduced separable fit at $N_f=5$.}
    \label{fig:lyap_asym_sep_5}
  \end{subfigure}\hfill
  \begin{subfigure}[t]{0.48\textwidth}
    \centering
    \includegraphics[width=\textwidth]{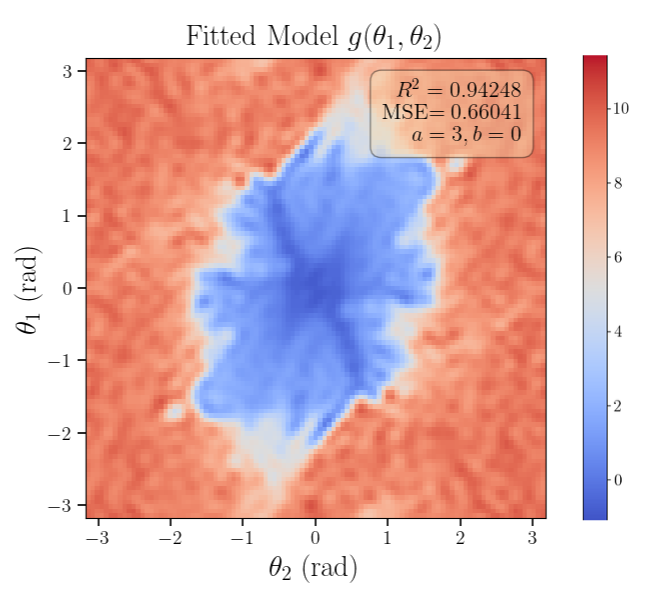}
    \caption{Parity-reduced separable fit at $N_f=20$.}
    \label{fig:lyap_asym_sep_20}
  \end{subfigure}

  \caption{Reconstruction quality for the asymmetric double-pendulum configuration. The separable tensor-product basis again provides a substantially better approximation than the phase-combination basis.}
  \label{fig:lyapunov_asym_comparison}
\end{figure}

The same qualitative pattern persists: the phase-combination basis remains limited, whereas the separable parity-reduced model stays highly effective. This confirms that the advantage of the separable construction is not specific to the symmetric baseline case. Figure~\ref{fig:lyapunov_asym_comparison} illustrates the reconstruction quality in the asymmetric setting.

\subsubsection{Interpretation}

The double-pendulum benchmark separates two aspects of the method. The dimensional prefactor fixes the overall scaling, but the accuracy is controlled by the dictionary used for the dimensionless modulation. Phase-combination harmonic features saturate early because they do not match the anisotropic structure of the Lyapunov field, whereas separable tensor-product dictionaries resolve the angular directions independently.

Symmetry reduction gives an additional reduction in feature count, but its effect is secondary to the choice of dictionary. On the symmetric benchmark, the phase-combination model plateaus near $R^2 \approx 0.82$, whereas the parity-reduced separable model reaches $R^2 = 0.95$.

\begin{figure}[htbp]
  \centering
  \includegraphics[width=1\textwidth]{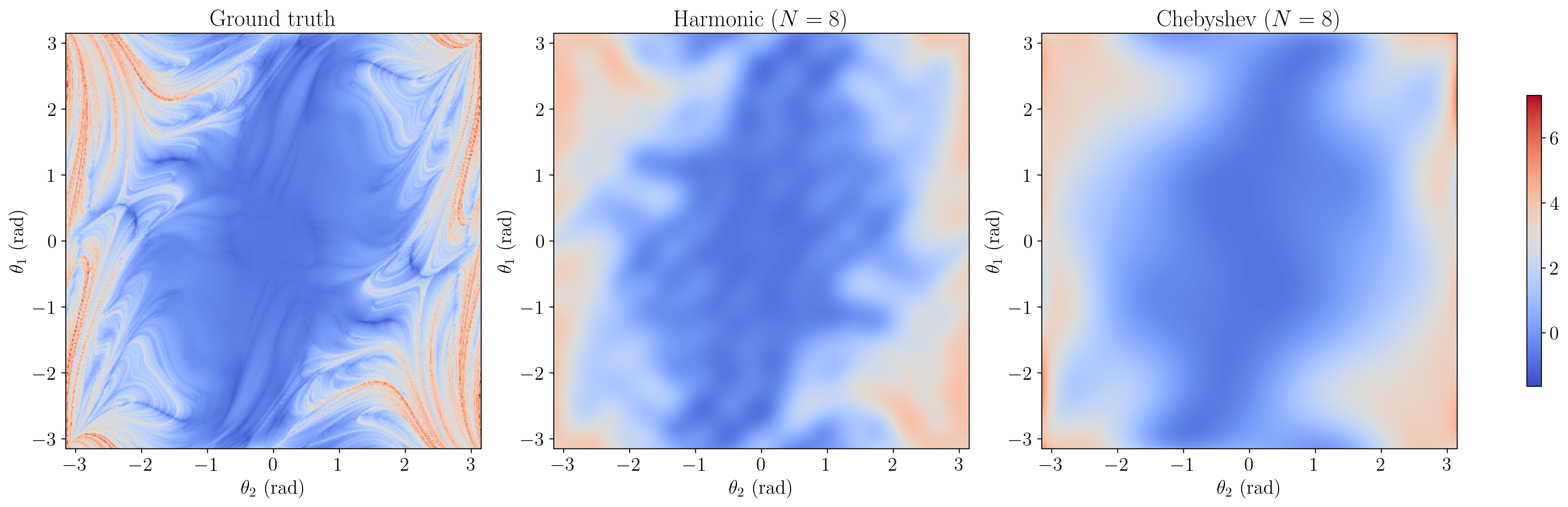}
  \caption{Comparison of separable harmonic and separable Chebyshev dictionaries on the double-pendulum Lyapunov field. Left: ground truth. Center: harmonic fit at $N_f=8$. Right: Chebyshev fit at $N_f=8$. The harmonic basis captures both marginal and interaction angular structure more faithfully.}
  \label{fig:cheb_dp}
\end{figure}

{
\subsubsection{Harmonic versus Chebyshev basis}

We also compare the separable harmonic dictionary against a separable Chebyshev polynomial basis on the same angular domain (Figure~\ref{fig:cheb_dp}). Chebyshev polynomials are orthogonal on $[-1,1]$, whereas the angular coordinates $\theta_1,\theta_2$ are naturally periodic on $[-\pi,\pi]$; mapping the angles to $[-1,1]$ introduces an artificial discontinuity at the domain boundaries. At matched truncation order $N_f=8$, the harmonic model achieves $R^2=0.882$ against $R^2=0.848$ for Chebyshev, and the gap persists with increasing $N_f$ ($R^2=0.907$ vs.\ $0.869$ at $N_f=12$). This is consistent with the angular nature of the invariant coordinates in this benchmark.
}

\begin{figure}[htbp]
  \centering
  \includegraphics[width=1\textwidth]{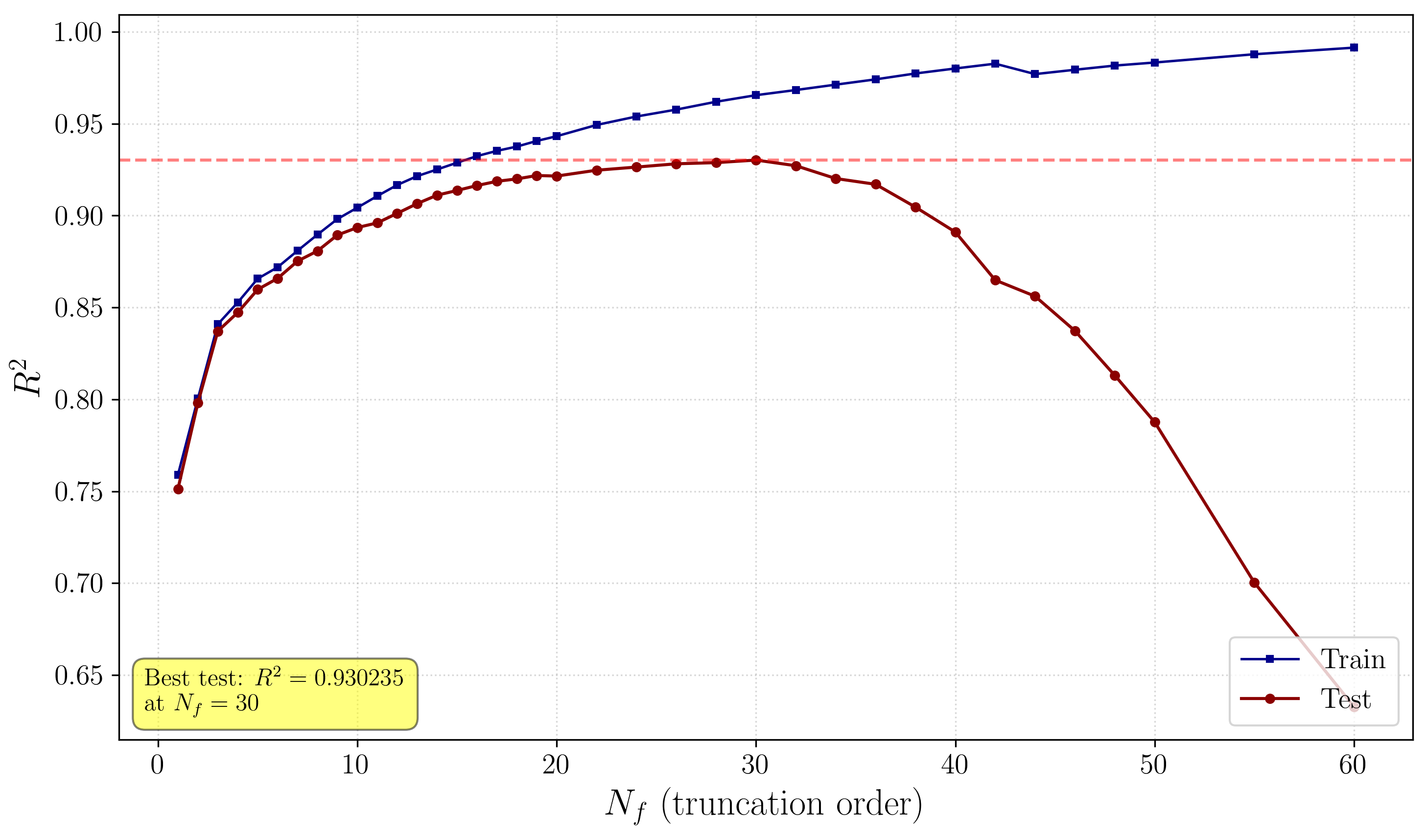}
  \caption{Train and test $R^2$ as a function of the truncation order $N_f$ for the parity-reduced separable model on an 80/20 split of 8000 subsampled grid points. Test performance peaks at $N_f\approx30$ and decreases at higher $N_f$ due to overfitting, indicating that the plateau reported in the earlier tables is a consequence of finite training samples rather than a limitation of the harmonic representation.}
  \label{fig:dp_convergence}
\end{figure}

{
\subsubsection{Convergence with truncation order}

The results reported in Tables~\ref{tab:lyapunov_phase}--\ref{tab:lyapunov_parity} are limited to three values of $N_f\in\{5,20,35\}$. To assess whether $R^2\approx0.95$ at $N_f=35$ represents saturation or a transient plateau, we extend the scan to $N_f=60$ using the parity-reduced separable model with a train/test split (Figure~\ref{fig:dp_convergence}). The training $R^2$ increases monotonically with $N_f$, reaching $0.991$ at $N_f=60$, as expected from a model with growing expressivity. The out-of-sample $R^2$, however, peaks at $N_f\approx30$ ($R^2=0.930$) and then declines to $0.633$ at $N_f=60$ due to overfitting with the available number of training samples. The observed plateau in the earlier tables is therefore not a property of the harmonic basis but rather reflects the finite sample budget: the model saturates when the number of features approaches the effective number of training degrees of freedom. This behavior is consistent with standard bias--variance trade-off expectations for regularized regression.}

\begin{figure}[htpb]
  \centering
  \includegraphics[width=1\textwidth]{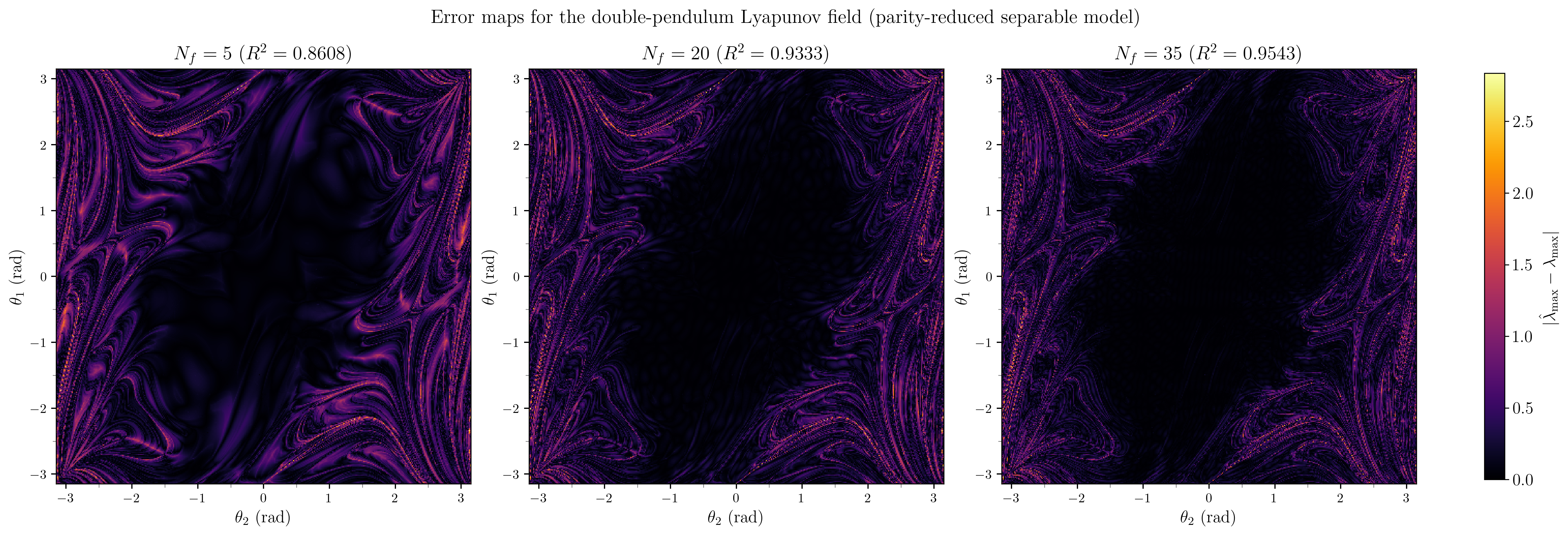}
  \caption{Heatmaps of the absolute prediction error $|\hat\lambda_{\max}-\lambda_{\max}|$ for the parity-reduced separable model at three truncation orders. Errors concentrate at the boundaries between regular and chaotic regions and narrow progressively with increasing $N_f$, but the interface structure persists.}
  \label{fig:dp_error}
\end{figure}

{
\subsubsection{Spatial distribution of the prediction error}

The aggregate metrics in Tables~\ref{tab:lyapunov_phase}--\ref{tab:lyapunov_parity} abstract away the spatial structure of the error. Figure~\ref{fig:dp_error} shows heatmaps of $|\hat\lambda_{\max}-\lambda_{\max}|$ over the angular domain for the parity-reduced separable model at three truncation orders. The error is systematically concentrated along the boundaries between regular and chaotic regions of the phase space, where the Lyapunov field exhibits sharp gradients. At $N_f=5$, errors of order $\mathcal{O}(1)$ span large portions of the domain; at $N_f=20$, the high-error regions shrink to thin filaments along the separatrix; at $N_f=35$, errors are further reduced in magnitude but remain localized at the same interface structures. This confirms that the primary challenge for the harmonic approximation is not uniform field complexity but the resolution of narrow transition zones, for which higher-order harmonic terms provide diminishing returns once the dominant interface width is captured.
}

\section{Discussion}
\label{sec:discussion}

The three benchmarks probe complementary aspects of the proposed framework: a
low-dimensional law largely determined by dimensional analysis, a
one-dimensional but transcendental modulation with a nontrivial prefactor
family, and a structured two-dimensional response surface with partial chaos.
Taken together, they show that dimensional consistency is useful, but that the
residual dimensionless dependence still has to be represented appropriately. The unconstrained pendulum
baseline in Section~\ref{sec:pendulum_unified} quantifies the sample-efficiency
gain: by fixing the exponent from dimensional analysis, the proposed model
avoids the ill-conditioned joint optimization over $(\alpha,w_0,w_1)$ and
reaches high accuracy with far fewer data points.

A first observation is that the dimensional component is more stable than the functional one. In the black-body benchmark, for example, the correct prefactor is recovered even in regimes where the harmonic approximation begins to overfit noise. This suggests that the admissible prefactor family acts as a low-complexity constraint, whereas the fitted dimensionless modulation carries the higher-variance part of the approximation.

A second observation is that dictionary geometry becomes important as soon as the invariant space has dimension greater than one. In the double-pendulum benchmark, phase-combination harmonic features saturate at comparatively modest accuracy, whereas separable tensor-product dictionaries yield a large improvement. The reason is geometric rather than purely spectral: the Lyapunov field is anisotropic, and a basis that couples all directions through a single phase is not well adapted to that structure. By contrast, separable dictionaries expose marginal and interaction effects directly and allow the representation to follow the directional organization of the target surface.

The experiments also show that symmetry information can be incorporated in a useful way. In the double-pendulum case, imposing joint parity reduces the number of fitted coefficients without degrading accuracy, and in the best models it improves efficiency at essentially no cost in expressivity on the physically relevant subspace. This illustrates the complementary roles of dimensional analysis and geometric symmetry: the former constrains the admissible scaling structure, while the latter restricts the dimensionless modulation within invariant space.

A further methodological point is that prefactor selection and functional approximation cannot be treated as fully independent. When the harmonic dictionary is too rigid, residual approximation error may leak into the prefactor scan, leading to small but systematic distortions of the recovered dimensional scaling. The black-body benchmark makes this interaction particularly clear. Conversely, when the dictionary is sufficiently expressive, the scan stabilizes near the physically correct prefactor. Prefactor selection should therefore be viewed not merely as a dimensional bookkeeping device, but as part of the overall bias-variance trade-off of the model.

Several limitations should also be kept in mind. First, although the present
work is primarily based on synthetic benchmarks, the COBE/FIRAS validation in
Section~\ref{sec:blackbody_results} provides a real-data check: the bootstrap
prefactor scan remains compatible with the expected scaling, but also shows the
limited statistical resolution of such a small measured dataset.

Overall, dimensional analysis defines a compact physically admissible hypothesis space. The experiments also show that, once multiple invariants are present, the learning stage depends strongly on matching the dictionary to the geometry of the response surface.

Regarding the choice of basis, harmonic dictionaries work well for the
transcendental and periodic invariant domains studied here. The Chebyshev
comparisons in Sections~\ref{sec:blackbody_results} and
\ref{sec:doublependulum_results} are useful ablations, but they do not outperform
the harmonic construction on these benchmarks.

\section{Conclusion}
\label{sec:conclusion}

We have introduced a data-driven method for learning physical responses under explicit dimensional constraints. Starting from the dimension matrix of the variables, the method constructs admissible dimensional prefactors and independent Buckingham invariants, and then learns the remaining dimensionless dependence through truncated harmonic expansions on invariant space. Once the dictionary is fixed, the fitting problem is linear in the coefficients and can be regularized by standard methods.

The numerical experiments cover three qualitatively different settings. It recovers the expected pendulum scaling,
identifies the correct black-body dimensional skeleton even under approximation
error and noise, and shows that separable tensor-product dictionaries with
symmetry reduction are essential for the double-pendulum Lyapunov field. The
convergence and error-map analyses further indicate that the remaining error is
mostly a finite-sample issue concentrated near regular-to-chaotic transition
regions.

More broadly, the study highlights three methodological points. First, dimensional consistency is a useful inductive bias for data-driven law discovery. Second, the separation between dimensional scaling and dimensionless modulation suggests a path to extensions in which the dimensional skeleton is fixed by prior knowledge while the modulation is refined as more data become available. Third, once the invariant space has dimension greater than one, the representation chosen for the dimensionless modulation becomes a central design decision.

The present work focuses on problems with low-dimensional invariant spaces, where prefactor scans and moderately rich harmonic dictionaries remain tractable. Extending the framework to larger invariant dimensions will require more structured approximation strategies, such as sparse truncations, adaptive dictionaries, or low-rank tensor constructions. The results obtained here suggest that combining dimensional analysis with explicit spectral approximation is a practical route toward learning physically admissible laws from data.

\section*{Data availability}

The datasets generated and/or analysed during the current study are available from the corresponding author on reasonable request. This includes the raw synthetic datasets generated for the controlled benchmarks and the processed data used to produce the figures and tables. The COBE/FIRAS spectrum used for the real-data validation is a public dataset originally reported by Fixsen et al.~\cite{fixsen1996}; the processed values used in the present analysis are also available from the corresponding author on reasonable request.

\section*{Funding}
{
This work was supported by the Cátedra Fundación ASISA--UEM de
Ciencias de la Salud, under internal project code P2025-15CA.
}

\section*{Author Contributions}
\textbf{Ernest Tarrus:} Writing - original draft, validation, software, methodology, investigation, formal analysis, and conceptualization. \textbf{Hector Gisbert:} Writing - original draft, visualization, supervision, methodology, investigation, formal analysis, and conceptualization.

\newpage

\begin{appendices}

\section{Differential form of unit equivariance}
\label{app:pde_proof}

For $\lambda=(\lambda_1,\dots,\lambda_R)\in(\mathbb{R}_{>0})^R$, introduce logarithmic coordinates
\begin{align}
t_r=\log \lambda_r,
\qquad r=1,\dots,R.
\end{align}
Under the induced change of units, each input variable transforms as
\begin{equation}
x_j(t)=x_j\exp\!\left(\sum_{r=1}^R a_{rj}t_r\right),
\qquad j=1,\dots,N.
\end{equation}
The unit-equivariance condition \eqref{eq:unit-equivariance} can then be written as
\begin{equation}
f\bigl(x_1(t),\dots,x_N(t)\bigr)
=
\exp\!\left(\sum_{r=1}^R b_r t_r\right)\,f(x_1,\dots,x_N).
\end{equation}
Differentiating with respect to $t_r$ at $t=0$ gives
\begin{equation}
\sum_{j=1}^{N}
\frac{\partial f}{\partial x_j}
\frac{\partial x_j}{\partial t_r}\bigg|_{t=0}
=
b_r\,f(x_1,\dots,x_N).
\end{equation}
Since
\begin{align}
\frac{\partial x_j}{\partial t_r}\bigg|_{t=0}=a_{rj}x_j,
\end{align}
we obtain the system
\begin{equation}
\sum_{j=1}^{N} a_{rj}\,x_j\,\frac{\partial f}{\partial x_j}
=
b_r\,f,
\qquad r=1,\dots,R.
\label{eq:app_euler_pde}
\end{equation}
This is the differential form of dimensional homogeneity. For a monomial ansatz
\begin{align}
f(\mathbf{x})=\prod_{j=1}^{N}x_j^{\beta_j},
\end{align}
substituting into \eqref{eq:app_euler_pde} yields
\begin{align}
\sum_{j=1}^{N} a_{rj}\beta_j=b_r,
\qquad r=1,\dots,R,
\end{align}
that is, the algebraic system $A\bm{\beta}=\mathbf{b}$. Thus, for monomial prefactors, the differential and algebraic formulations of unit equivariance are equivalent.

\newpage

\section{Summary of black-body radiation with noise}\label{app:noisePlanck}

\begin{figure}[htbp]
  \centering
  \includegraphics[width=1\textwidth]{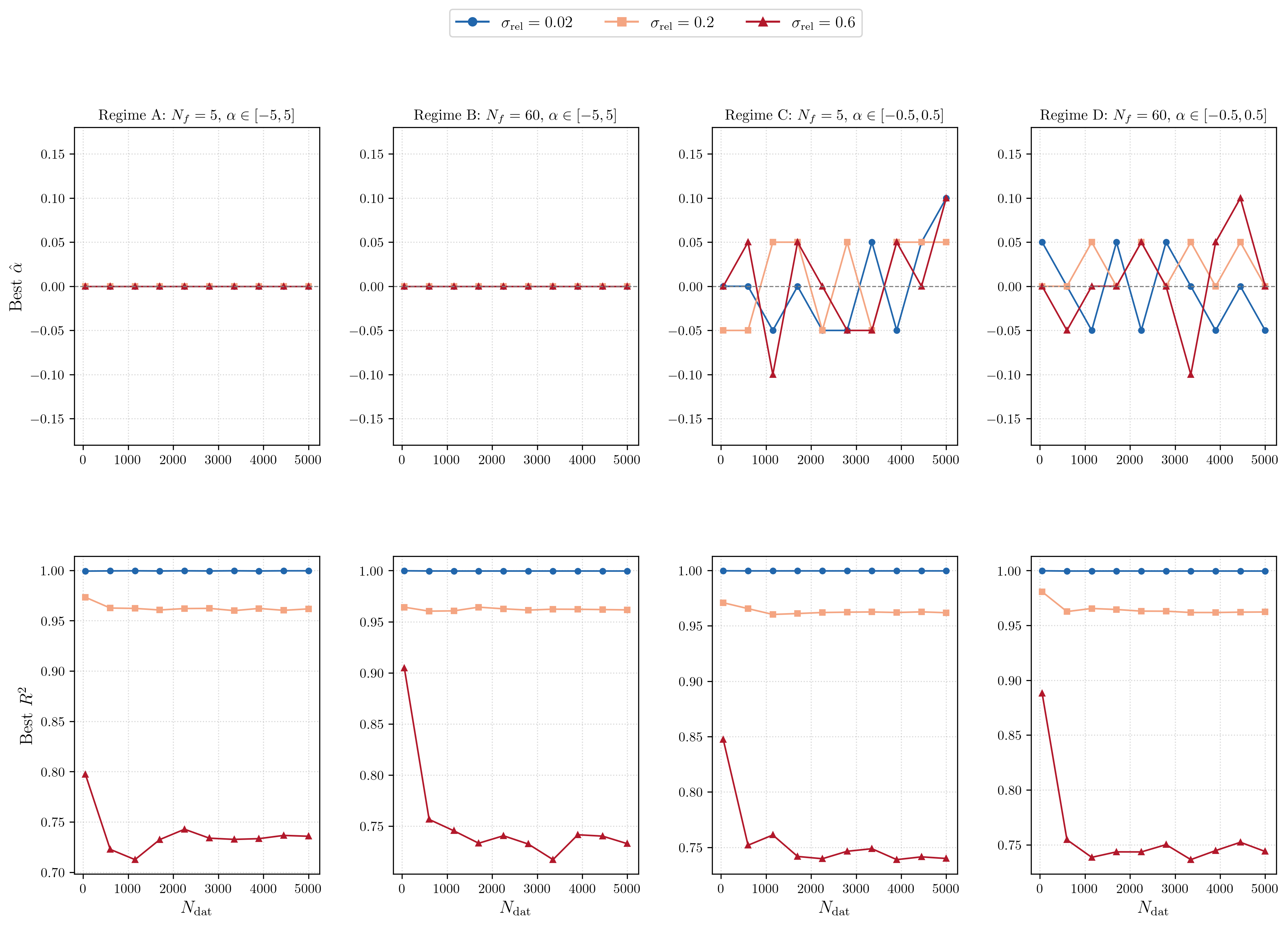}
  \caption{Summary of the noisy black-body experiments across the four regimes. Top: recovered prefactor exponent $\hat\alpha$ vs.\ $N_{\mathrm{dat}}$. Bottom: best $R^2$ vs.\ $N_{\mathrm{dat}}$. Three noise levels are shown per panel. The prefactor exponent remains robust across all regimes, while $R^2$ degrades with noise and, for large-$N_f$ models, also with small $N_{\mathrm{dat}}$.}
  \label{fig:bb_noise_summary}
\end{figure}

{The full numerical results of the four noisy regimes described in Section~\ref{sec:blackbody_results} are summarized in Figure~\ref{fig:bb_noise_summary}. The top row shows the recovered prefactor exponent $\hat\alpha$ as a function of the dataset size $N_{\mathrm{dat}}$ for each regime and noise level. Across all regimes, the method recovers values very close to the theoretical exponent $\alpha=0$, with deviations that stay within $\pm0.1$ except at the smallest sample sizes. The bottom row shows the corresponding out-of-sample $R^2$. For the low-noise regime ($\sigma_{\mathrm{rel}}=0.02$), $R^2$ remains above $0.999$ at all dataset sizes and both $N_f=5$ (Regime~A, C) and $N_f=60$ (Regime~B, D). At higher noise levels, the low-$N_f$ model maintains $R^2\approx0.96$ ($\sigma_{\mathrm{rel}}=0.2$) and $R^2\approx0.73$ ($\sigma_{\mathrm{rel}}=0.6$), while the high-$N_f$ model exhibits overfitting at small $N_{\mathrm{dat}}$ (e.g., $R^2=0.72$ at $N_{\mathrm{dat}}=3350$ for Regime~B, $\sigma_{\mathrm{rel}}=0.6$). These results confirm that the dimensional component (the prefactor exponent $\alpha$) is substantially more stable than the functional component ($R^2$ from the harmonic fit) and that the qualitative behavior is robust to the chosen scan range for $\alpha$.}

\clearpage

\end{appendices}

\bibliography{sn-bibliography}

\end{document}